\RequirePackage[l2tabu, orthodox]{nag}
\documentclass[12pt]{article}

\usepackage[parfill]{parskip}
\usepackage[margin=1in]{geometry}
\usepackage[defaultlines=3,all]{nowidow}
\usepackage[T1]{fontenc}
\usepackage{microtype}
\usepackage[english]{babel}
\usepackage{libertine}             % Libertine (serif), Biolinium (sans).
\usepackage{titlesec}
\usepackage{nicefrac}
\titleformat*{\section}{\Large\bfseries}

\usepackage{amsmath}
\usepackage{amssymb}
\usepackage{amsthm}
\usepackage{mathtools}

\usepackage[dvipsnames]{xcolor}

\usepackage[sort]{natbib}
\usepackage{hyperref}
\hypersetup{colorlinks,linktoc=all}
\usepackage[all]{hypcap}
\hypersetup{citecolor=orange}
\hypersetup{linkcolor=black}
\hypersetup{urlcolor=MidnightBlue}

\usepackage{wrapfig}
\usepackage{microtype}
\usepackage{graphicx}
\usepackage{nicefrac}
\usepackage{subfigure}
\usepackage{booktabs} % professional-quality tables
\usepackage{makecell}
\usepackage{multirow}
\usepackage{wrapfig}
\usepackage{caption}
\usepackage{subcaption}
\usepackage{enumitem}
\usepackage{color}

\usepackage{algorithm}
\usepackage{algorithmic}
\usepackage{xcolor}

\newtheorem{Theorem}{Theorem}

\newtheorem{Proposition}[Theorem]{Proposition}

\newcommand{\cS}{\mathcal{S}}
\newcommand{\cD}{\mathcal{D}}
\newcommand{\cA}{\mathcal{A}}

\newcommand{\cT}{\mathcal{T}}
\newcommand{\cZ}{\mathcal{Z}}
\newcommand{\cK}{\mathcal{K}}

\newcommand{\x}{\pmb x}
\newcommand{\y}{\pmb y}
\newcommand{\z}{\pmb z}
\newcommand{\K}{\mathbf{K}}
\newcommand{\X}{\mathbf{X}}
\newcommand{\bL}{\mathbf{L}}
\newcommand{\bH}{\mathbf{H}}

\usepackage{hyperref}
\hypersetup{colorlinks=true, linkcolor=orange, citecolor=orange, urlcolor=orange}
\usepackage[capitalize,noabbrev]{cleveref}
\usepackage{url}

\newcommand{\E}{\mathbb{E}}

\newcommand{\alphab}{\pmb \alpha}

\newcommand{\onevec}{\pmb 1}

\newcommand{\bG}{\mathbf{G}}

\newcommand{\bGamma}{\mathbf{\Gamma}}
\newcommand{\bR}{\mathbf{R}}

\title{Amortized Linear-time Exact Shapley Value 
       for Product-Kernel Methods}

\author{
    Majid Mohammadi\textsuperscript{\rm 1,2}\thanks{Denotes joint first authors} \and 
    Siu Lun Chau\textsuperscript{\rm 3}\footnotemark[1] \and 
    Krikamol Muandet\textsuperscript{\rm 2} 
}
\date{
\footnotesize{
    \textsuperscript{\rm 1}Rational Intelligence Lab, CISPA Helmholtz Center for Information Security, Germany\\
    \textsuperscript{\rm 2}Department of Computer Science, Vrije Universiteit Amsterdam, The Netherlands\\
    \textsuperscript{\rm 3}Epistemic Intelligence \& Computation Lab, College of Computing \& Data Science, Nanyang Technological University, Singapore\\[2ex]}
}

\begin{document}

\maketitle

\begin{abstract}

Kernel methods are widely used in machine learning and statistics for their flexibility and expressive power, yet their black-box nature limits adoption in high-stakes applications. Shapley value--based attribution methods such as SHAP, and kernel-specific adaptations including RKHS-SHAP, provide a principled framework for explainability---but exact computation of Shapley values is generally intractable, forcing existing approaches to rely on approximations that incur unavoidable estimation error. We introduce PKeX-Shapley, an algorithm that exploits the multiplicative structure of product kernels to compute exact Shapley values for all $d$ features in quadratic time in $d$. The method rests on a distribution-free removal operator intrinsic to the product-kernel structure: removing a feature replaces its kernel factor with the multiplicative identity. This yields a parameter-free value function---requiring no sampling and no density estimation---and uniquely determines a functional decomposition of the model. Building on this value function, we develop shared recursive formulations that evaluate all feature attributions jointly, achieving amortized linear time per feature with numerical stability. Beyond predictive modeling, the framework extends to widely used kernel-based discrepancies such as the Maximum Mean Discrepancy (MMD) and the Hilbert--Schmidt Independence Criterion (HSIC), providing new tools for interpretable statistical analysis.
\end{abstract}


\newpage 

%\begin{links}
%\link{Code}{https://anonymous.4open.science/r/FGPX-Shapley-L-CF0A/}
%\end{links} 

%%%%%%%%%%%%%%%%%%%%%%%%%%%%%%%%%%%%%%%
%%%%%%%%%%%%% Introduction
%%%%%%%%%%%%%%%%%%%%%%%%%%%%%%%%%%%%%%%
% \vspace{-1em}
\section{Introduction}

Shapley values~\citep{shapley}, a solution concept from cooperative game theory, provide a principled axiomatic framework for feature attribution in machine learning (ML)~\citep{shap, sage}. Their rigorous foundation has driven widespread adoption in the explainable AI community~\citep{many_sv}: a model's output---predictions, losses, or other quantities of interest---can be \emph{fairly} distributed across input features according to their individual and joint contributions. A range of algorithms has been developed to estimate Shapley values under different modelling assumptions, from model-agnostic methods such as Kernel SHAP~\citep{shap} to model-specific methods that leverage structural properties for statistical or computational gains: TreeSHAP for tree-based models~\citep{treeshap}, GPSHAP for Gaussian processes~\citep{gp_shap}, Deep SHAP for deep neural networks~\citep{shap}, and, most relevant to our work, RKHS-SHAP for kernel methods~\citep{rkhs_shap}. Kernel methods are notable well beyond prediction: they underpin a broad range of statistical inference tasks, including measuring distributional closeness~\citep{gretton2006kernel, naslidnykkernel}, two-sample testing~\citep{schrab2025unified, chau2025credal}, goodness-of-fit testing~\citep{Chwialkowski16:GOF, Liub16:KSD}, independence testing~\citep{hsic_test}, and causal inference and discovery~\citep{kernel_causal1, zhang2026instrumental}. As kernel methods see growing use in high-stakes applications for their flexibility and expressive power, interpretability has become essential.

Despite their rigorous game-theoretic foundation, Shapley values face two major challenges when applied to ML. The first is to define and estimate a value function that quantifies the contribution of a feature coalition. Ideally, this function should capture the model's behavior when the complementary features are absent. A natural approach retrains the model on each subset and uses the resulting prediction as the value~\citep{sv_retainingModel}, but the exponential number of retrainings makes this infeasible. A more practical alternative simulates feature absence via marginal or conditional expectations of the model's output given the retained subset; see \citet{many_sv} for further options. \citet{rkhs_shap} pursue this route in the context of kernel methods, exploiting the structure of reproducing kernel Hilbert spaces (RKHS) and using kernel distributional embeddings~\citep{muandet2017kernel} to estimate value functions nonparametrically, thereby avoiding density estimation. An increasingly popular alternative defines the value function through functional decomposition~\citep{gevaert2024unifying,coop_funcdecompos,mohammadi2025exact}, identifying the contribution of a feature subset with the component of the model that depends on those features.
% Despite their rigorous game-theoretic foundation, Shapley values face two major challenges when applied to machine learning. The first is to define and estimate a suitable value function that quantifies the contribution of a coalition of features. Ideally, this function should capture the model’s behavior when the complementary features are absent. A natural approach is to retrain a model on each subset and use the resulting prediction as the value~\citep{sv_retainingModel}, but this is computationally infeasible as it requires an exponential number of retrainings. A more common alternative simulates feature absence through marginal or conditional expectations of the model’s output, given the fixed subset; see \citet{many_sv} for other options. \citet{rkhs_shap} tackle this in the context of kernel methods by exploiting the structure of reproducing kernel Hilbert spaces (RKHS) and using kernel distributional embeddings~\citep{muandet2017kernel} to estimate value functions nonparametrically, thereby avoiding density estimation and providing reliable computation. Another increasingly popular perspective defines the value function through functional decompositions~\citep{gevaert2024unifying,coop_funcdecompos,mohammadi2025exact}, where the contribution of a feature subset is identified with the component of the model that depends on the corresponding subset. 
% \textcolor{red}{Alan: Maybe here we can already mention functional decomposition is also a way that is gaining popularity as a value function.}

\begin{figure}
    \centering
    \includegraphics[width=0.75\linewidth]{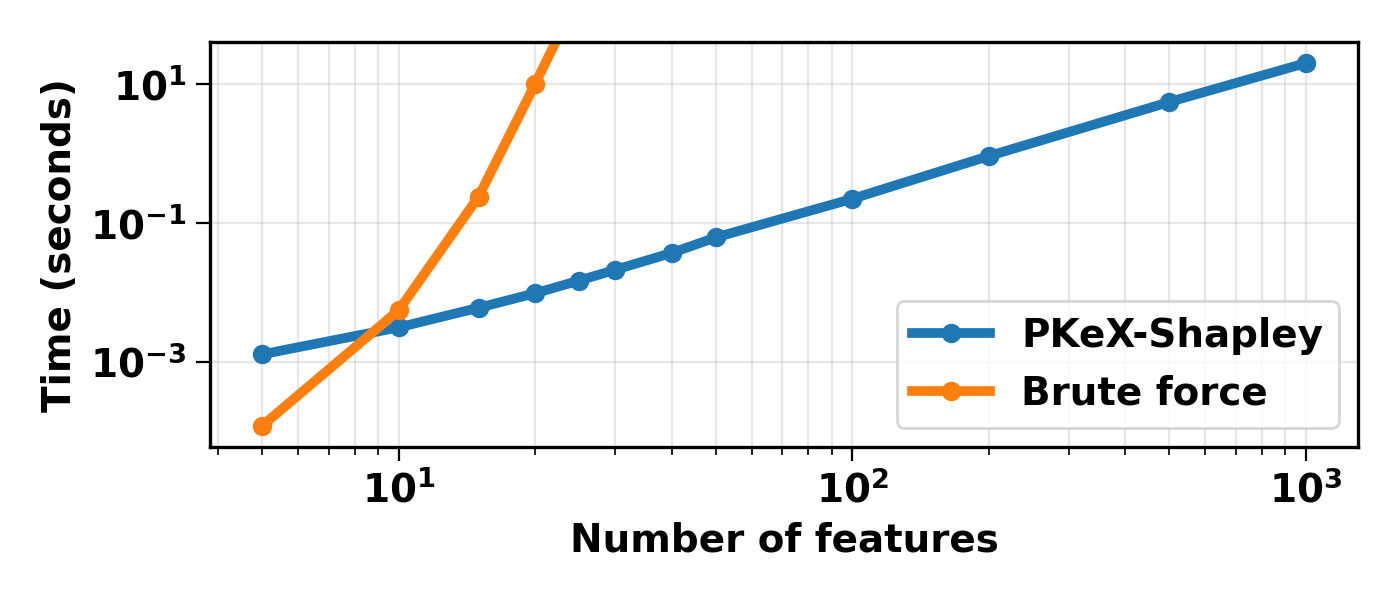}
    \caption{Execution time vs.\ number of features (log–log). Brute-force reaches at most 20 features in 20 s; PKeX-Shapley scales to 1,000.}
    \label{fig:time_cmp_bf}
\end{figure}

Once a value function is chosen, the second challenge is computing the Shapley values themselves. Exact evaluation requires the value function on all $2^d$ feature subsets, which quickly becomes prohibitive. Approximation methods---Monte Carlo sampling, regression-based estimators---sidestep the exponential cost by evaluating only a subset of coalitions, but introduce estimation errors that grow with dimensionality~\citep{shap_issues}. Model-specific structure can sometimes be exploited to recover exactness: TreeSHAP, for instance, leverages tree decompositions for exact polynomial-time computation. RKHS-SHAP improves the statistical estimation of the value function but still relies on regression-based approximation for the Shapley values, inheriting the same computational limitations.

% Once a value function is chosen, the second challenge lies in efficiently estimating the Shapley values themselves.  Naive exact computation requires evaluating the value function over all $2^d$ possible feature subsets for $d$ features, which is computationally demanding. To alleviate this, approximation techniques such as Monte Carlo sampling and regression-based methods are widely used, relying on a smaller set of evaluations. While these approaches reduce computational costs, they inevitably introduce estimation errors that grow with feature dimensionality~\citep{shap_issues}. In some cases, model-specific structures can be exploited for efficiency---for example, TreeSHAP leverages tree decompositions for exact polynomial-time computation. By contrast, RKHS-SHAP improves the statistical estimation of the value function but still relies on regression-based approximations for the Shapley values themselves, and thus inherits the same computational limitations.

To design an efficient Shapley value--based attribution algorithm for kernel methods, we focus on the subclass employing product kernels---which we call \emph{product-kernel methods}---and introduce \emph{PKeX-Shapley} (\textbf{P}roduct-\textbf{K}ernel-based \textbf{eX}act \textbf{Shapley} attribution). Product kernels are simple to implement yet retain strong theoretical guarantees: the product of universal kernels is itself universal, so the resulting RKHS can approximate any continuous function whenever each component kernel is sufficiently expressive~\citep{szabo2018characteristic}. This narrows the scope relative to~\citet{rkhs_shap}, which considers generic kernels, but the additional structure lets us address both challenges raised above directly. We first identify the \emph{removal operator} intrinsic to product-kernel models: removing a feature $j$ amounts to replacing its kernel factor by the multiplicative identity, $k_j \mapsto 1$. This choice preserves the product structure, requires no background distribution, and induces a parameter-free value function---no sampling, no density estimation. By \citet{gevaert2024unifying}, every removal operator determines a unique \emph{functional decomposition}; ours admits a simple closed form indexed by feature subsets. Building on this value function, we derive a recursive formulation that computes the Shapley value of a single feature in quadratic time. Organizing these recursions across features via a prefix--suffix construction of elementary symmetric polynomials, we then show that all $d$ Shapley values can be computed jointly in $O(d^2 n)$ time---an \emph{amortized linear-time} cost per feature in $d$~\citep[Chapter 17]{clrs}; see Figure~\ref{fig:time_cmp_bf} for comparison against brute-force computation.

As an additional, independent contribution, we demonstrate that our framework extends beyond kernel-based predictive models. In particular, we show how our method applies to popular kernel-based statistical discrepancies, including the Maximum Mean Discrepancy (MMD) and Hilbert–Schmidt Independence Criterion (HSIC), thereby enabling interpretable statistical inference. A Python implementation of our method is publicly available~\citep{pkex_shapley_code}.

\section{Background}

\paragraph{Notation.} Let $\cD$ denote the set of $d$ features, and $2^\cD$ its power set. The training set $\{(\x^{(i)}, y^{(i)})\}_{i=1}^n$ consists of $n$ samples, where $\x^{(i)} \in \mathbb{R}^d$ and $y^{(i)} \in \mathbb{R}$ (or a discrete label set for classification). Let $\X \in \mathbb{R}^{n \times d}$ be the feature matrix, and $\X_{\cS}\in\mathbb{R}^{n\times s}$ the submatrix restricted to features in subset $\cS \subseteq \cD$, and we write $\X_j := \X_{\{j\}}$. We use capital letters for random variables, bold capital letters for matrices, calligraphic letters for sets, and bold lowercase letters for vectors. The restriction of a vector $\x$ to features in $\cS$ is denoted by $\x_\cS$. The element-wise product and division are respectively denoted by $\odot$ and $\oslash$, and expectation by $\mathbb{E}$. A symmetric positive (semi-)definite kernel function over $\cD$ is denoted by $k$, and its restriction to a subset $\cS$ by $k_\cS$. The corresponding kernel matrices are denoted by $\K$ and $\K_\cS$, respectively. All proofs are provided in Appendix \ref{apx:proof}. 

\paragraph{Shapley value.} The Shapley value~\citep{shapley} is a widely used game-theoretic solution concept for feature attribution in predictive models. It assigns importance to each input feature by averaging its marginal contribution across all feature subsets, weighted by coalition size. Given a value function $v:2^\mathcal{D} \to \mathbb{R}$ that quantifies the contribution of a feature subset, the Shapley value for feature $j$ is
\begin{equation}
    \phi_j = \sum_{\cS \subseteq \cD \setminus \{j\}} \mu(|\cS|) \big( v(\cS\cup \{j\}) - v(\cS) \big),
\end{equation}
where $\mu(s) = \nicefrac{s!\,(d - s -1)!}{d!}$ and $v(\cS \cup \{j\}) - v(\cS)$ is the marginal contribution of $j$ to $\cS$. For any fixed value function, the Shapley value is the unique attribution satisfying efficiency, null player, symmetry, and linearity~\citep{shapley}. In explainability, however, the choice of $v$ is itself a non-trivial design decision. Several formulations have been proposed~\citep{many_sv}; two are dominant. The \emph{interventional} value function~\citep{interventional_shap},
$$
v_{\x}(\cS) = \E_{X_{\cD \setminus \cS}}\!\left[f(\x_{\cS},X_{\cD \setminus \cS})\right],
$$
replaces missing features with samples from their marginal distributions, while the \emph{observational} value function~\citep{chen2020true},
$$
v_{\x}(\cS) = \E\!\left[f(X) \,\middle|\, X_{\cS}=\x_{\cS}\right],
$$
conditions on the observed features. Both require sampling from marginal or conditional distributions, incurring repeated model evaluations and a dependence on background samples that can substantially shape the resulting explanations~\citep[Chapter 21]{shap_book}.

\paragraph{Functional decomposition.} An increasingly popular alternative defines the value function through a functional decomposition of the learned model~\citep{gevaert2024unifying, coop_funcdecompos, mohammadi2025exact}, expressing $f$ as a sum of components indexed by feature subsets, $$f(\x)=\sum_{\cS\subseteq \cD} f_{\cS}(\x_{\cS}),$$
where each $f_{\cS}$ depends only on the variables in $\cS$. The value function can then be defined as the Möbius transform of the functional components $\{f_S\}$ and thus $$\nu_{\x}(S) = \sum_{\cT \subseteq S}f_{\cT}(\x_{\cT}),$$ aggregating all sub-coalition effects of $S$. 

This framing connects Shapley-based attribution to a century of statistical theory: ANOVA~\citep{st1989analysis}, Sobol indices~\citep{owen_anova,owen_sobol}, partial dependence~\citep{berzal2002relational}, and generalized additive models~\citep{hastie2017generalized} are all functional decomposition methods that long predate modern XAI. The advantage is not merely historical. Writing $f$ as $\sum_\cS f_\cS$ makes a \emph{structural} claim about the model that can be inspected componentwise, and many desirable properties of attribution methods, analogues of Shapley's axioms stated directly in terms of $f$ rather than an induced cooperative game, admit clean characterizations at the decomposition level~\citep[Section 6.3]{gevaert2024unifying}.

Furthermore, \citet{gevaert2024unifying} sharpen this picture by establishing a correspondence between functional decompositions and \emph{removal operators}---mappings that render the model independent of a given feature subset. Every removal operator induces a unique decomposition, and every decomposition arises this way. Designing a value function therefore reduces to a single modelling choice: pick a sensible removal operator and read off the decomposition. We adopt the removal-operator perspective. In the next section, we show that the natural removal operator for product-kernel models---replacing each absent kernel factor with the multiplicative identity---yields a parameter-free value function and induces a functional decomposition with a simple closed form.
%%%%%%%%%%%%%%%%%%%%%%%%%%%%%%%%%%%%%%%%%%%%%%%%%%%%%%%%%%%%
%%%%%%%%%%%% Exact Shapley Value for Product Kernels
%%%%%%%%%%%%%%%%%%%%%%%%%%%%%%%%%%%%%%%%%%%%%%%%%%%%%%%%%%%%
% \vspace{-0.5em}
\section{Exact Shapley Value for Product-Kernel Learning Methods}\label{sec: mainmethod}
% \vspace{-0.5em}
% This section derives an algorithm for computing all $d$ Shapley values in amortized linear time in $d$ per feature--quadratic time in $d$ overall--for product-kernel methods, including support vector machines and kernel ridge regression. The value function is induced by a natural, distribution-free removal operator and admits a closed form requiring no sampling or density estimation.

This section develops an algorithm that computes all $d$ Shapley values for product-kernel methods---including support vector machines and kernel ridge regression---in quadratic time in $d$ overall, or amortized linear time per feature. The value function is induced by a natural, distribution-free removal operator and admits a closed form, requiring neither sampling nor density estimation.

%\vspace{-0.5em}
\subsection{Value Function for Product-Kernel Methods}

Product-kernel methods rely on kernel functions to capture complex relationships between input features and output. A kernel-based decision function is generally expressed as
$$f(\x) = \sum_{i=1}^{n} \alpha_i k(\x, \x^{(i)}) = \alphab^\top k(\X, \x)$$
where \( k \) is a kernel function and \( \alpha_i \) are the model-specific coefficients. 
When $k$ is a product kernel, it can be expressed as:
%\begin{align}\label{eq:product kernel}
$$
k(\x, \x^{(i)}) =  \prod_{j \in \cD} k_j(x_j, x^{(i)}_j).
$$
%\end{align}
Product-kernel methods are widely used in ML for their simplicity and effectiveness in modelling similarities in high-dimensional data~\citep{product_kernel1}. They also come with strong theoretical guarantees: if the base kernels are universal---capable of approximating any continuous function on the marginal input---then the product kernel inherits this property~\citep{szabo2018characteristic}. Several widely used kernels are of product form, including the radial basis function (RBF) with isotropic or anisotropic bandwidth,
$$
k(\x, \x') = \exp\!\left(-\|\x - \x'\|^2/2\sigma^2\right) = \prod_{j=1}^d \exp\!\left(-(x_j - x'_j)^2/2\sigma^2\right);
$$
further examples are given in Appendix~\ref{apx:product kernels}. For product-kernel methods, we utilize the canonical neutral-factor removal: removing a feature subset $\cT$ corresponds to replacing each kernel factor $k_j$, $j\in\cT$, with the multiplicative identity $1$, giving
\[
P_\cT(f)(\x) = \alphab^\top k_{\cD\setminus\cT}(\X_{\cD\setminus\cT},\x_{\cD\setminus\cT}).
\]
This removal preserves the product-kernel structure and requires no background distribution, sampling, or density estimation---absent features contribute a neutral factor. By the canonical additive decomposition framework of \citet{gevaert2024unifying}, once this removal family is chosen, it uniquely determines the corresponding functional decomposition via M\"{o}bius inversion on the subset lattice. The resulting components are given in Proposition~\ref{prop:func-decomp}.

% ------- Proposition 1: functional decomposition -------
\begin{Proposition}\label{prop:func-decomp}
The removal operator above induces the functional decomposition $f = \sum_{\cS\subseteq\cD} f_\cS$ with components
$$
f_{\cS}(\x_{\cS}) = \sum_{i=1}^{n} \alpha_i \prod_{j\in\cS}\bigl(k_j(x_j,x^{(i)}_j)-1\bigr), 
$$ where $f_\emptyset = \sum_{i=1}^n \alpha_i$.
\end{Proposition}
% \vspace{-1em}
The value function for coalition $\cS$ is obtained by applying the removal operator to the absent features.
% The value function for a coalition $\cS$ is then obtained by applying this removal operator to the absent features~\citep{gevaert2024unifying},
% \[v_{\x}(\cS) := P_{\cD\setminus\cS}(f)(\x) - P_{\cD}(f)(\x),\]
% where the fully-removed baseline $P_{\cD}(f)(\x) = f_\emptyset = \sum_{i=1}^n\alpha_i$ is subtracted to ensure $v_{\x}(\emptyset)=0$. Since $f_\emptyset$ is a constant independent of $\x_\cS$, it cancels in every marginal contribution $v_{\x}(\cS\cup\{j\})-v_{\x}(\cS)$ and has no effect on the Shapley values. We therefore drop it henceforth.

% ------- Proposition 2: value function closed form -------
\begin{Proposition}\label{prop:func-decomp-value}
The value function induced by the canonical neutral-factor removal is
\begin{equation}\label{eq:func-decomp-value}
v_{\x}(\cS) \;=\; \alphab^\top k_{\cS}(\X_{\cS}, \x_{\cS}) - f_\emptyset,
\end{equation}
where $f_\emptyset = \sum_{i=1}^n \alpha_i$, satisfying $v_{\x}(\emptyset) = 0$ and $v_{\x}(\cD) = f(\x) - f_\emptyset$.
\end{Proposition}
Since $f_\emptyset$ is constant, it cancels in every marginal contribution $v_{\x}(\cS\cup\{j\})-v_{\x}(\cS)$ and has no effect on the Shapley values. We therefore drop it henceforth and work with $v_{\x}(\cS) = \alphab^\top k_{\cS}(\X_{\cS}, \x_{\cS})$, which is exact and parameter-free: it requires no sampling and no density estimation, and can be evaluated in $O(|\cS|\cdot n)$ time. Given this value function, we next derive an efficient algorithm to compute exact Shapley values.

%%%%%%%%%%%%%%%%%%%%%%%%%%%%
% \vspace{-0.5em}
\subsection{Exact Shapley Values Computation}
Building on the value function above, we now show that the Shapley value of any feature admits a recursive form. This recursion underlies the computational gains of our approach and provides the pathway to exact computation.

% Building on the value function above, we demonstrate how Shapley values associated with any feature can be written in a recursive form. This recursion underlies the computational gains of our approach and provides an algorithmic pathway for exact computation.

\begin{Theorem}\label{th:shapley recursive}
Let $\cZ := \{\z_1,\ldots ,\z_d\}$ with
$
\z_i := k_i(\X_i,x_i)
$, $\cZ_{-j} = \cZ \setminus \{\z_j\}$,
and $e_q(\cZ)$ the elementary symmetric polynomials (ESPs) of order $q$ over $\cZ$,
defined as
\[
e_q(\cZ_{-j}) = \frac{1}{q} \sum_{r=1}^{q} (-1)^{r-1} e_{q-r}(\cZ_{-j}) \odot p_r(\cZ_{-j}),
\]
where \(p_r (\cZ) = \sum_{\z_i \in \cZ} \z_i^r\) is the element-wise degree-$r$ power sum.
For product-kernel learning methods with coefficients $\alphab$, the Shapley value $\phi_j^{\x}$
for feature $j$ of instance $\x$ under the functional decomposition value function can then be expressed as
{\small
\begin{align}\label{eq:shapley ng}
\phi_j^{\x}= \alphab^\top \big( \big( \z_j - \onevec \big) \bigodot
\sum_{q=0}^{d-1} \mu(q) \, e_q\big(\cZ_{-j}\big)  \big).
\end{align}
}
\end{Theorem}

% \vspace{-0.5em}
The recursion follows from Newton's identities for ESPs~\citep{sym_poly_book}; we defer the details to Appendix~\ref{apx:newton identity}. Intuitively, under the functional decomposition value function, the multiplicative structure of the product kernel allows us to factorize the marginal contribution $v_{\x}(\cS\cup\{j\})-v_{\x}(\cS)$ as $\alphab^\top\!\big((k_j(\X_j,x_j)-\onevec)\odot k_{\cS}(\X_{\cS},\x_{\cS})\big)$. This factorization lets us pull the sum over $\cS\subseteq\cD\setminus\{j\}$ inside the inner product with $\alphab$ and express it directly in terms of weighted ESPs $e_q(\cZ_{-j})$, where $\z_i = k_i(\X_i, x_i)$.
% These ESPs admit a recursive computation via Newton’s identities, leading to an $O(d^2)$ algorithm for computing a single exact Shapley value. A similar idea has been used in the context of additive Gaussian processes~\citep{agp}, but to our knowledge, it has not previously been leveraged for computing Shapley values under a functional decomposition value function in amortized linear time.

% Next, we show the additivity of the explanation for the value function in \eqref{eq:kernel value function} (see Appendix \ref{apx:additivity} for discussion on the additivity of explanations for different values for the null game). 

% \begin{Lemma}\label{lem:learning additivity}
% For any instance $\x$, the sum of Shapley values satisfies:
% $\sum_{j=1}^d \phi_j^{\x} = f(\x) - f_\emptyset(\x)$
% where \( f_\emptyset(\x) = \sum_{i=1}^n \alpha_i \) represents the baseline contribution with no features.
% \end{Lemma}

\paragraph{From one to all Shapley values.}
The recursion in Theorem \ref{th:shapley recursive} shows that a \emph{single} Shapley value $\phi_j^{\x}$ can be computed in $O(d^2)$ time, since each $e_q$ follows from the power sums $p_r(\cZ_{-j})$ via Newton's identities. Applying this procedure independently to every $j \in \cD$, however, ignores substantial overlap between the sets $\cZ_{-j}$: the ESPs $e_q(\cZ_{-j})$ across different $j$ share many intermediate quantities, which can be reused.
% The recursive formulation in \Cref{th:shapley recursive} shows that computing a \emph{single} Shapley value $\phi_j^{\x}$ is quadratic in $d$, since each $e_q$ can be obtained via Newton’s identities from the power sums $p_r(\cZ_{-j})$. However, repeating this procedure independently for all $j \in \cD$ ignores the substantial overlap between the sets $\cZ_{-j}$: the ESPs $e_q(\cZ_{-j})$ for different $j$ share many intermediate terms. In principle, one can exploit this structure and reuse ESP computations across features.
% so that \emph{all} Shapley values ${\phi_j^{\x}}_{j\in\cD}$ can be obtained in overall $O(d^2)$ time, rather than $O(d^3)$.

To this end, we develop a numerically stable method that exploits structural properties of ESPs to compute the sum $\sum_q \mu(q) e_q(\cZ_{-j})$ from equation~\eqref{eq:shapley ng}. The key tool is the generating-polynomial representation of ESPs, which organizes interaction terms by degree and makes intermediate quantities reusable across features. For any ordered index set $\cA \subseteq \cD$, define
$$
p_{\cA}(s) := \bigodot_{i\in \cA} \big(\onevec + \z_i\, s\big),
$$
where $s$ is a univariate variable. Expanding the product gives
$$
p_{\cA}(s) = \sum_{q=0}^{|\cA|} e_q(\cA)\, s^q,
$$
so the coefficient of $s^q$ is the degree-$q$ ESP over the vectors indexed by $\cA$. For a fixed index $j\in\cD$, we construct the ESPs over $\cZ_{-j}$ in~\eqref{eq:shapley ng} by splitting the remaining vectors into a \emph{prefix} (indices before $j$) and a \emph{suffix} (indices after $j$):
\begin{align*}
p^{\mathrm{pref}}_{j-1}(s) := \bigodot_{i=1}^{j-1} \big(\onevec + \z_i\, s\big) = \sum_{t=0}^{j-1} \bGamma_{j-1,t}\, s^t, \qquad
p^{\mathrm{suf}}_{j+1}(s) := \bigodot_{i=j+1}^{d} \big(\onevec + \z_i\, s\big) = \sum_{u=0}^{d-j} \bR_{j+1,u}\, s^u,
\end{align*}
where $\bGamma_{j-1,t}$ and $\bR_{j+1,u}$ are the element-wise ESPs of degrees $t$ and $u$ over the vectors preceding and following index $j$, respectively. The leave-one-out polynomial then factorizes as
$
p_{-j}(s) = p^{\mathrm{pref}}_{j-1}(s) \odot p^{\mathrm{suf}}_{j+1}(s),
$
involving only vectors in $\cZ_{-j}$ by construction.

Shapley value computation thus reduces to evaluating the inner sum of weighted degree-$q$ ESPs. The following proposition shows that prefix–suffix polynomial constructions enable the reuse of intermediate ESP quantities across features, yielding efficient computation of all Shapley values.

\begin{Proposition}
\label{prop:omega_prefix_suffix}
Let $\bGamma_{j-1,t}$ and $\bR_{j+1,u}$ be as above, and $\bG_{j+1,t} := \sum_{u=0}^{d-j} \mu(t+u)\, \bR_{j+1,u}$. Then
$$
\sum_{q=0}^{d-1} \mu(q)\, e_q(\cZ_{-j})
=
\sum_{t=0}^{j-1} \bGamma_{j-1,t} \odot \bG_{j+1,t}.
$$
\end{Proposition}

% \vspace{-0.5em}

% Proposition~\ref{prop:omega_prefix_suffix} shows that the efficient computation of all Shapley values reduces to constructing the prefix coefficients $\bGamma_{j,t}$ and the aggregated suffix quantities $\bG_{j,u}$ for $\forall j\in \cD, t < j, u \leq d-j$: The Shapley value of each feature could be computed in linear-time if we have these two quantities. Next, we show that both quantities admit simple recursive constructions and can be computed in quadratic time in $d$.
\textbf{Time and memory complexity.} Proposition~\ref{prop:omega_prefix_suffix} reduces the computation of all Shapley values to constructing the prefix coefficients $\bGamma_{j,t}$ and the aggregated suffix quantities $\bG_{j,u}$ for $j\in \cD$, $t < j$, $u \leq d-j$: given these, each Shapley value can be obtained in linear time. We now show that both quantities admit simple recursive constructions and can themselves be computed in $O(d^2)$ time.
\begin{Proposition}
\label{prop:time_memory_complexity}
The prefix coefficients $\bGamma=\{\bGamma_{j,t}\}$ and aggregated suffix quantities $\bG=\{\bG_{j,t}\}$ can be computed and stored in $O(d^2 n)$ time and $O(d^2 n)$ memory. %The construction relies exclusively on element-wise additions and multiplications of vectors.
\end{Proposition}
Proposition~\ref{prop:time_memory_complexity} shows that the prefix and suffix ESP coefficients can be computed in $O(d^2 n)$ time, after which each Shapley value $\phi_j^{\x}$ is obtained in $O(dn)$ time. \textbf{Computing all $d$ Shapley values therefore costs $O(d^2 n)$ in total, or $O(dn)$ per feature on average.}
% Proposition~\ref{prop:time_memory_complexity} shows that we can compute the prefix and suffix ESP coefficients in $O(d^2 n)$ time, after which each individual Shapley value $\phi_j^{\x}$ is obtained in $O(dn)$ time. Thus, the joint computation of all $d$ Shapley values requires
% $O(d^2 n)$ time, corresponding to an average cost of $O(dn)$ per
% feature.

%Thus, the overall complexity of computing all $d$ Shapley values jointly is $O(d^2 n)$, corresponding to an \emph{amortized linear-time}~\citep{clrs} cost per feature in the number of features $d$.

\paragraph{Comparison to TreeSHAP.} A value function with similar multiplicative structure arises in TreeSHAP, where interpolation-based algorithms recover leave-one-out ESPs via polynomial division~\citep{linear_treeshap,q-shap}. These methods work well when the polynomial degree is bounded by the tree depth $D \ll d$, giving Linear TreeSHAP an $\mathcal{O}(DL)$ per-tree complexity. For product-kernel methods, however, the polynomial degree is $d-1$, and interpolation becomes numerically unstable already at moderate dimensions (e.g., $d \approx 50$); see Appendix~\ref{apx:interpolation}. PKeX-Shapley avoids this by relying only on element-wise additions and multiplications, remaining stable at large $d$ with $\mathcal{O}(d^2 n)$ total cost. Notably, applying Linear TreeSHAP to kernel methods would view the model as an ensemble of $n$ trees---one per training point, each with depth and leaf count $d$---also yielding $\mathcal{O}(d^2 n)$ overall, but at the cost of the numerical instability above.

\section{Explaining kernel-based statistical discrepancies}

We extend the removal-operator framework to two widely used kernel-based statistical discrepancies: the \emph{Maximum Mean Discrepancy} (MMD) and the \emph{Hilbert--Schmidt Independence Criterion} (HSIC). The same principle applies in both cases: removing a feature replaces its kernel factor with the multiplicative identity, yielding a parameter-free value function in closed form. We give the corresponding recursive Shapley formulations based on Newton's identities; the prefix--suffix construction transfers directly and is omitted for brevity.

% We extend the removal-operator framework to two widely used kernel-based statistical discrepancy measures: the \textit{Maximum Mean Discrepancy} (MMD) and the \textit{Hilbert-Schmidt Independence Criterion} (HSIC). In both cases the same principle applies: removing a feature corresponds to replacing its kernel factor with the multiplicative identity $1$, yielding a parameter-free value function with an explicit closed form. We present the corresponding recursive Shapley value formulations based on Newton’s identities; for brevity we omit the prefix–suffix stable construction, but it is straightforwardly applicable to both measures.

%%%%%%%%%%%%%%%%%%%%%%%%%%%%%%%%%%%%%%%%%%%%%%%%%%%%%%%%%%%%
%%%%%%%%%%%% MMD Explaining
%%%%%%%%%%%%%%%%%%%%%%%%%%%%%%%%%%%%%%%%%%%%%%%%%%%%%%%%%%%%
\subsection{Distributing the discrepancy: Explaining MMD}

The MMD quantifies the difference between two probability distributions in terms of their kernel mean embeddings: $$\text{MMD}^2(\mathbb{P},\mathbb{Q}) := \|\mu_{\mathbb{P}}-\mu_{\mathbb{Q}}\|_{\mathcal{H}_k}^2$$ where $\mathcal{H}_k$ is the RKHS associated with the kernel $k$ and $\mu_{\mathbb{P}} := \int k(\x,\cdot)\,d\mathbb{P}(\x) \in \mathcal{H}_k$ is the kernel mean embedding of $\mathbb{P}$~\citep[Sec. 3.5]{muandet2017kernel}. The embedding $\mu_{\mathbb{Q}}$ is defined analogously.
Based on the samples \( \{\x^{(i)}\}_{i=1}^{n} \sim \mathbb{P} \) and \( \{\z^{(i)}\}_{i=1}^{m} \sim \mathbb{Q} \), the empirical estimate of MMD, expressed entirely in terms of $k$, is given by 
{\small
\begin{equation*}
    \widehat{\text{MMD}}^2(\mathbb{P}, \mathbb{Q}) = \frac{1}{n(n-1)} \sum_{i \neq j} k(\x^{(i)}, \x^{(j)}) + \frac{1}{m(m-1)} \sum_{i \neq j} k(\z^{(i)}, \z^{(j)}) - \frac{2}{nm} \sum_{i, j} k(\x^{(i)}, \z^{(j)}).
\end{equation*}
}
In this section, we propose an attribution method that allocates the overall discrepancy measured by MMD among the variables involved. Such attributions are useful in many settings, including explaining MMD-based test statistics in hypothesis testing~\citep{mmd_hypo} and identifying the variables driving covariate shift~\citep{mmd_covar_shift}. For a product kernel $k$, removing a feature subset $\cT$ replaces each factor $k_q$, $q\in\cT$, with $1$, leaving only the sub-kernel $k_{\cD\setminus\cT}$. Applied to MMD, this removal induces a natural, parameter-free value function, and the corresponding functional decomposition is uniquely determined.
% In this section, we propose an attribution method to allocate the overall discrepancy measured by MMD among the involved variables. Such an attribution is useful in many problems, including explaining MMD-based statistics for hypothesis testing (e.g., \citet{mmd_hypo}) and determining the contribution of variables to covariance shift (e.g., \citet{mmd_covar_shift}). For a product kernel $k$, removing a feature subset $\cT$ means replacing each factor $k_q$, $q\in\cT$, with $1$, so that only the sub-kernel $k_{\cD\setminus\cT}$ remains. Applying this removal to MMD yields a natural, parameter-free value function, and the corresponding functional decomposition is uniquely determined by this removal operator.

\begin{Theorem}\label{th:mmd func decomp}
For product kernel MMD, the removal operator $k_q \mapsto 1$ for $q \in \cT$ induces the following:
\begin{itemize}[leftmargin=*]
    \item[(i)] The value function representing the contribution of the variables in \(\cS\) is
    % {\small
    % \begin{equation}\label{eq:mmd value}
        $$
        v_{\text{MMD}}(\cS) = \frac{1}{n(n-1)} \sum_{i \neq j} k_{\cS}(\x^{(i)}_{\cS}, \x^{(j)}_{\cS}) + \frac{1}{m(m-1)} \sum_{i \neq j} k_{\cS}(\z^{(i)}_{\cS}, \z^{(j)}_{\cS}) - \frac{2}{nm} \sum_{i, j} k_{\cS}(\x^{(i)}_{\cS}, \z^{(j)}_{\cS}).
        $$
    % \end{equation}
    % }
    \item[(ii)] The MMD function admits the unique functional decomposition into subset contributions induced by this removal operator:
    {\small
        \begin{align*}
        &\widehat{\text{MMD}}^2(\mathbb{P}, \mathbb{Q}) = \sum_{\cS \subseteq \cD} \bigg[ \frac{1}{n(n-1)} \sum_{i \neq j} \prod_{q \in \cS} \big( k_q(x^{(i)}_q, x^{(j)}_q) - 1 \big) \cr
        & \qquad + \frac{1}{m(m-1)} \sum_{i \neq j} \prod_{q \in \cS} \big( k_q(z^{(i)}_q, z^{(j)}_q) - 1 \big)
        - \frac{2}{nm} \sum_{i, j} \prod_{q \in \cS} \big( k_q(x^{(i)}_q, z^{(j)}_q) - 1 \big) \bigg].
        \end{align*}
    }
\end{itemize}
\end{Theorem}
Applying the same construction as in Theorem~\ref{th:shapley recursive} to $v_{\mathrm{MMD}}$ yields an analogous recursive formulation. For the first term in $v_{\mathrm{MMD}}$, the multiplicative structure of the product kernel gives
$$
v_{\mathrm{MMD}}(\cS \cup \{q\}) - v_{\mathrm{MMD}}(\cS) 
= k_{\cS \cup \{q\}}(\x^{(i)}_{\cS \cup \{q\}}, \x^{(j)}_{\cS \cup \{q\}}) - k_{\cS}(\x^{(i)}_{\cS}, \x^{(j)}_{\cS}) \\
= \big(k_{q}(x^{(i)}_{q}, x^{(j)}_{q}) - 1\big)\, k_{\cS}(\x^{(i)}_{\cS}, \x^{(j)}_{\cS}).
$$
Pulling $k_{q}(x^{(i)}_{q}, x^{(j)}_{q}) - 1$ outside the summation in the Shapley value lets us express the total sum over $\cS \subseteq \cD \setminus \{q\}$ as a weighted ESP. The following proposition formalizes this.

% By applying the same trick as in Theorem \ref{th:shapley recursive} to the value function $v_{\text{MMD}}$, we can obtain a similar recursive formulation of Shapley values for MMD. That is, for the first term in $v_{\text{MMD}}$, we use the multiplicative structure of product kernels and write $v_{\text{MMD}}(\cS \cup \{q\}) - v_{\text{MMD}}(\cS)$ as $k_{\cS \cup \{ q \} }(\x^{(i)}_{\cS \cup \{ q \}}, \x^{(j)}_{\cS \cup \{ q \}}) - k_{\cS}(\x^{(i)}_{\cS}, \x^{(j)}_{\cS}) = \big(k_{q}(x^{(i)}_{q}, x^{(j)}_{q})-1\big)k_{\cS}(\x^{(i)}_{\cS}, \x^{(j)}_{\cS})$. By pushing out $k_{q}(x^{(i)}_{q}, x^{(j)}_{q})-1$ from the summation in the Shapley value, we can express the total sum over $\cS \subseteq \cD \setminus \{q\}$ as the weighted ESPs. The following proposition summarizes this. 

\begin{Proposition}\label{prop:mmd sv}
Let $ \cZ^{(\x, \x')} = \{ k_1(x_1, x'_1), \ldots, k_d(x_d, x'_d) \}$, and $ e_r(\cZ_{-q}^{(\x,\x')}) $ determined as
{\small\[
% $
e_r(\cZ_{-q}^{(\x,\x')}) = \frac{1}{r} \sum_{s=1}^{r} (-1)^{s-1} e_{r-s}(\cZ_{-q}^{(\x,\x')}) \, p_s(\cZ_{-q}^{(\x,\x')}),
% $
\]
}
where $ p_s(\cZ) = \sum_{z \in \cZ} z^s $ represents the degree-$s$ power sum.
Further, let $\gamma_q(\x,\x')$ be defined as $$\gamma_q(\x,\x') := ( k_q(x_q, x'_q) - 1 ) \sum_{r=0}^{d-1} \mu(r) e_r(\cZ_{-q}^{(\x, \x')}).$$
Then, for product kernels, the Shapley value for the MMD can be recursively computed as
{\small
\begin{equation*}
    \phi_q^{\text{MMD}} = \tfrac{1}{n(n-1)} \sum_{i \neq j} \gamma_q(\x^{(i)},\x^{(j)}) + \tfrac{1}{m(m-1)} \sum_{i \neq j} \gamma_q(\z^{(i)},\z^{(j)}) - \tfrac{2}{nm} \sum_{i, j} \gamma_q(\x^{(i)},\z^{(j)}).
\end{equation*}
}
%{\small 
%\begin{align}
%    \phi_q^{\text{MMD}} &= \tfrac{1}{n(n-1)} \sum_{i \neq j} \bigg( \big( k_q(x^{(i)}_q, x^{(j)}_q) - 1 \big) \sum_{r=0}^{d-1} \mu(r) e_r(\cZ_{-q}^{(\x^{(i)}, \x^{(j)})}) \bigg) \cr
%    &\quad + \tfrac{1}{m(m-1)} \sum_{i \neq j} \bigg( \big(k_q(z^{(i)}_q, z^{(j)}_q) - 1 \big) \sum_{r=0}^{d-1} \mu(r) e_r(\cZ_{-q}^{(\z^{(i)}, \z^{(j)})}) \bigg) 
%    \cr &\quad - \tfrac{2}{nm} \sum_{i, j} \bigg( \big(k_q(x^{(i)}_q, z^{(j)}_q) - 1 \big) \sum_{r=0}^{d-1} \mu(r) e_r(\cZ_{-q}^{(\x^{(i)}, \z^{(j)})}) \bigg).
%\end{align}
%}
\end{Proposition}

The Shapley values $\phi_q^{\text{MMD}}$ allow us to allocate the overall distributional discrepancy between $\mathbb{P}$ and $\mathbb{Q}$ across the variables, identifying the most influential ones in distinguishing the two distributions.

%%%%%%%%%%%%%%%%%%%%%%%%%%%%%%%%%%%%%%%%%%%%%%%%%%%%%%%%%%%%
%%%%%%%%%%%% HSIC for feature Selection
%%%%%%%%%%%%%%%%%%%%%%%%%%%%%%%%%%%%%%%%%%%%%%%%%%%%%%%%%%%%
\subsection{Distributing the dependence: Explaining HSIC}

The HSIC is a kernel-based dependence measure between two random variables. Let $X$ and $Y$ be random variables equipped with reproducing product kernels $k(\cdot,\cdot)$ and $l(\cdot,\cdot)$, respectively. Then
$$
\mathrm{HSIC}(X,Y) := \|\mathcal{C}_{XY}\|^2_{\mathrm{HS}},
$$
where $\mathcal{C}_{XY}$ is the cross-covariance operator and $\|\cdot\|_{\mathrm{HS}}$ is the Hilbert--Schmidt (HS) norm; see~\citet[Sec.~3.6]{muandet2017kernel} for technical details. Given a sample $\{(\x^{(i)}, \y^{(i)})\}_{i=1}^{n} \sim \mathbb{P}(X,Y)$, HSIC admits the empirical estimator
$$
\widehat{\mathrm{HSIC}}(X, Y) = \frac{1}{(n-1)^{2}}\, \mathrm{tr}(\K\bH\bL\bH),
$$
where $\K, \bL \in \mathbb{R}^{n\times n}$ are the kernel matrices with $\K_{ij} = k(\x^{(i)},\x^{(j)})$ and $\bL_{ij} = l(\y^{(i)},\y^{(j)})$, and $\bH = I - \tfrac{1}{n}\onevec\onevec^\top$ is the centreing matrix. 

Computing $\mathrm{HSIC}(X,Y)$ gives the overall dependence between $X$ and $Y$. We now consider the setting where $X$ is a feature vector $\x$ and $Y$ is a scalar outcome $y$, and ask how the overall dependence can be distributed across the features of $\x$. This is particularly useful for feature selection and global sensitivity analysis, where the target is typically univariate.

% The HSIC is a kernel-based dependence measure that quantifies the statistical dependence between two random variables. Let $X$ and $Y$ be two random variables with $k(\cdot,\cdot)$ and $l(\cdot,\cdot)$ as reproducing (product) kernels defined on them. Then, $\text{HSIC}(X,Y) := \|\mathcal{C}_{XY}\|^2_{\text{HS}}$, where $\mathcal{C}_{XY}$ is the cross-covariance operator and $\|\cdot\|_{\text{HS}}$ is the Hilbert-Schmidt (HS) norm; see, e.g.,~\citep[Sec. 3.6]{muandet2017kernel} for technical details.
% Given a sample $\{(\x^{(i)}, \y^{(i)})\}_{i=1}^{n} \sim \mathbb{P}(X,Y)$, $\text{HSIC}(X,Y)$ can be estimated as $\widehat{\text{HSIC}}(X, Y) = (n-1)^{-2} \mathrm{tr}(\K\bH\bL\bH)$
% where $\K \in \mathbb{R}^{n\times n}$ is the kernel matrix computed using the kernel $k$, i.e, $\K_{ij} = k(\x^{(i)},\x^{(j)})$, $\bL \in \mathbb{R}^{n\times n}$ is the kernel matrix computed using the kernel $l$, i.e, $\bL_{ij} = l(\y^{(i)},\y^{(j)})$, and $\bH = I - \frac{1}{n} \onevec \onevec^\top$ is the centering matrix ensuring zero mean in the feature space. 

% Computing $\text{HSIC}(X,Y)$ gives us the overall dependence between $X$ and $Y$. We now focus on scenarios where $X$ represents a random variable of the feature vector $\x$ and $Y$ represents the scalar prediction outcome $y$. Our interest is then to distribute the overall dependence over the features in $\x$. This is particularly useful for feature selection and global sensitivity analysis, where the target is usually univariate. 

For a product kernel $k$ on $X$, the kernel matrix factorizes element-wise as $\K = \bigodot_{j\in\cD} \K_j$, where $(\K_j)_{ab} = k_j(x^{(a)}_j, x^{(b)}_j)$. Removing feature $j$ corresponds to replacing $\K_j$ with $\onevec\onevec^\top$, the all-ones matrix, which is the element-wise multiplicative identity ($\mathbf{A} \odot \onevec\onevec^\top = \mathbf{A}$) and the kernel matrix for the constant kernel $k_j \equiv 1$. Applied to HSIC, this yields a natural, parameter-free value function, and the corresponding functional decomposition is uniquely determined by this removal operator.

\begin{Theorem}\label{th:hsic func decomp}
For the HSIC with the product kernel on $X$, the removal operator $\K_j \mapsto \onevec\onevec^\top$ for $j \notin \cS$ induces the following:
\begin{itemize}[leftmargin=*]
    \item[(i)] The value function that quantifies the contribution of the variable set \(\cS\) to the overall dependence is
    % {\small
    % \begin{equation}
    % \label{eq:hsic value}
    $$
        v_{\text{HSIC}}(\cS) = {(n-1)^{-2}}\mathrm{tr} \big( \bH \bL \bH  \K_{\cS} \big).
    $$
    % \end{equation}
    % }
    \item[(ii)] The HSIC function admits the unique functional decomposition into subset contributions induced by this removal operator: $$\widehat{\text{HSIC}}(X,Y) =  \frac{1}{(n-1)^2}\sum_{\cS \subseteq \cD} \mathrm{tr} \bigg( \bH \bL \bH \bigodot_{j \in \cS} (\K_j - \onevec\onevec^\top) \bigg).$$
    % {\small
    % \begin{equation*}
    %     \widehat{\text{HSIC}}(X,Y) =  \frac{1}{(n-1)^2}\sum_{\cS \subseteq \cD} \mathrm{tr} \bigg( \bH \bL \bH \bigodot_{j \in \cS} (\K_j - \onevec\onevec^\top) \bigg).
    % \end{equation*}
    % }
\end{itemize}
\end{Theorem}

Applying the same trick as in Proposition \ref{prop:mmd sv} to the value function $v_{\text{HSIC}}$
% in Equation \eqref{eq:hsic value} 
yields a similar recursive formulation of Shapley values for HSIC.

\begin{Proposition}\label{prop:hsic sv}
For the product kernel, the Shapley value for HSIC can be recursively computed as: 
{\small
\[
\phi_j^{\text{HSIC}} =  \frac{1}{(n-1)^2}\mathrm{tr} \left( \bH \bL \bH \bigg( (\K_j - \onevec\onevec^\top) \bigodot \sum_{q=0}^{d-1} \mu(q) E_q(\cK_{-j}) \bigg) \right).
\]
}
where $\cK = \{\K_1, \dots, \K_d\}$ with $\K_i$ being the kernel matrix for feature $i$ only, and
$
E_q(\cK_{-j}) = \frac{1}{q} \sum_{r=1}^{q} (-1)^{r-1} E_{q-r}(\cK_{-j}) \bigodot P_r(\cK_{-j}),
$
% {\small 
% \[
% E_q(\cK_{-j}) = \frac{1}{q} \sum_{r=1}^{q} (-1)^{r-1} E_{q-r}(\cK_{-j}) \bigodot P_r(\cK_{-j}),
% \]
% }
is the ESPs with $P_r(\cK) = \sum_{\K_i \in \cK} \K_i^r$ being the element-wise degree-$r$ power sum.
\end{Proposition}

The Shapley values $\phi_j^{\text{HSIC}}$ allow us to allocate the overall dependence between $X$ and $Y$ across individual features fairly, identifying the most influential ones for prediction. Our results can be generalized to scenarios when both $X$ and $Y$ are multivariate. The attribution of overall dependence to the involved multivariate variables has other applications in statistical inference as well, e.g., kernel-based (conditional) independence testing~\citep{hsic_test, hsic_test2}.

\begin{figure*}[!t]
\centering\includegraphics[width=0.85\linewidth]{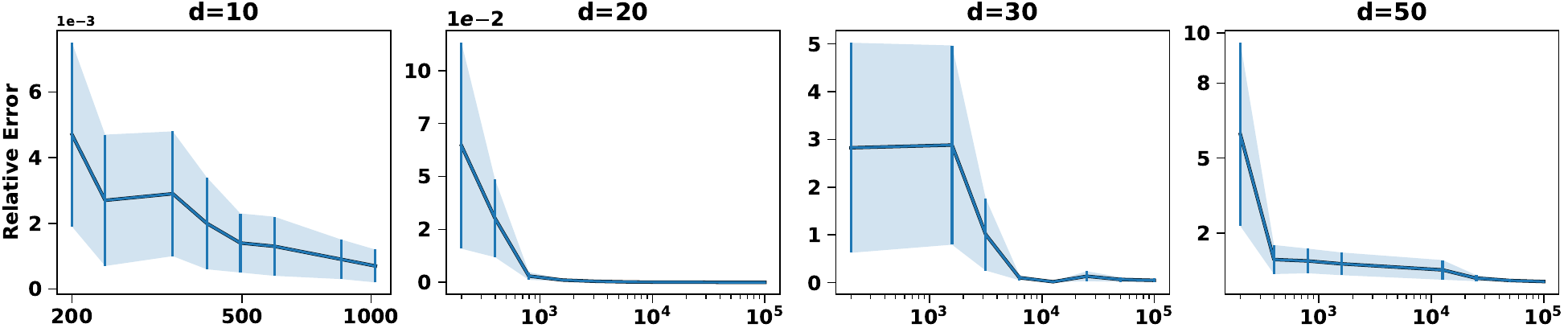}
    %\vspace{-0.7em}
    \caption{Relative estimation error of regression-based Shapley values versus the exact recursive values, shown across coalition sample sizes for feature dimensions $d=\{10,20,30,50\}$. }
        \label{fig:local rec reg}
        \vspace{-0.5em}
\end{figure*}

%%%%%%%%%%%%%%%%%%%%%%%%%%%%%%%%%%%%%%%%%%%%%%%%%%%%%%%%%%%%
%%%%%%%%%%%% Experiments
%%%%%%%%%%%%%%%%%%%%%%%%%%%%%%%%%%%%%%%%%%%%%%%%%%%%%%%%%%%%
\section{Experiments}
We demonstrate the effectiveness of PKeX-Shapley for product-kernel methods through a series of experiments conducted on a 24-core machine with 16GB RAM. We further provide the experimental setups, training procedure, and extra experiments in Appendix~\ref{apx:experiments}, including a feature selection case study using HSIC (Appendix~\ref{apx:hsic exp}).

\subsection{Effectiveness of Recursion in Local Explanations}\label{sec:exp 1 2}

\paragraph{Recursion vs. regression formulation.} 
We first demonstrate empirically that PKeX-Shapley outperforms sampling-based approximations such as Kernel SHAP. We generate four synthetic datasets, each with $1000$ samples and $d \in \{10, 20, 30, 50\}$ features drawn independently from a standard normal distribution. Labels follow a linear model with additive Gaussian noise ($\sigma = 0.1$). For each dataset, we train a support vector regression model with an RBF kernel and use it to compare explanation methods.
% First, we empirically demonstrate the advantage of PKeX-Shapley over sampling-based approximations such as Kernel SHAP. We generate four synthetic datasets, each with $1000$ samples and \(d\in\{10,20,30,50\}\) features. Features are independently drawn from a standard normal distribution, and labels are generated using a linear model with additive Gaussian noise ($\sigma=0.1$). For each dataset, we train a support vector regression model with an RBF kernel and use the trained model to compare explanation methods. %(see Appendix~\ref{apx:experimental setup} for training details).

% We first evaluate the performance of PKeX-Shapley on simulated regression tasks, highlighting both its runtime efficiency and the effectiveness of its value function in identifying informative features. 

% For fair comparison between recursion and regression methods, we employ the same value function in Definition~\ref{def:vS}. Specifically, we first use our polynomial-time recursive computation to obtain exact Shapley values \({\phi_j^{\x}}\). Then, we utilise the regression-based estimator analogous to Kernel SHAP but uses our value function in Definition~\ref{def:vS}. We employs the paired coalition sampling scheme~\citep{covert2021improving} of different sizes (from $200$ to $10^5$) and compute values \({\hat\phi_j}^{\x}\) by solving the regression problem in Kernel SHAP. To demonstrate the estimation error in the regression formulation, we compute the relative deviation error for 100 arbitrarily chosen instances, i.e., $\sum_{j=1}^d |\phi_j^{\x} - \hat\phi_j^{\x}|/\lvert\phi_j^{\x}\rvert.$

For a fair comparison, both methods use the same value function. We compute the exact Shapley values $\phi_j^{\x}$ with PKeX-Shapley, then approximate them using a regression-based estimator analogous to Kernel SHAP. The estimator follows the paired coalition sampling strategy of \citet{covert2021improving}, draws between $200$ and $10^5$ coalitions, and obtains $\hat{\phi}_j^{\x}$ by solving the resulting weighted linear regression. We report the average relative deviation
$\sum_{j=1}^d \lvert \phi_j^{\x} - \hat{\phi}_j^{\x} \rvert / \lvert \phi_j^{\x} \rvert$
over $100$ randomly selected instances. 

Figure~\ref{fig:local rec reg} plots the relative error against sample size for each dataset, averaged over the 100 instances. At $d = 10$, $200$ samples (roughly $20\%$ of all possible coalitions) already drive the error to about $0.005$. At $d = 50$, by contrast, $200$ samples leave the error above $9.0$; $10^4$ samples still leave it above $1.0$; and only at $10^5$ does it approach $0.05$. The required sample size grows sharply with dimensionality, making sampling-based methods unreliable in high-dimensional settings---and underscoring the value of fast exact computation.

\paragraph{Effectiveness of PKeX-Shapley.}
We evaluate the effectiveness of PKeX-Shapley through three \emph{synthetic} regression datasets with $n=1000$ samples in $\mathbb{R}^{50}$ where only the first one-third of the features (denoted by $\cS$, with $\lvert \cS \rvert = 17$) influence the response, and the remaining 33 features are redundant. The target functions over $\cS$ are: a degree-5 polynomial, a degree-10 polynomial, and a squared-exponential response of the form $y = \exp\!\big(-\sum_{i \in \cS} x_i^2\big)$. For each dataset, we train a support vector regressor with an RBF kernel and compute feature attributions using PKeX-Shapley alongside several baselines: RKHS-SHAP~\citep{rkhs_shap}, GEMFIX~\citep{gemfix}, BiSHAP~\citep{bivariateSHAP}, and Sampling SHAP~\citep{sampling_shap}, with 500 and 1000 coalition samples. All baseline methods use a fixed background set of 100 points for value-function estimation.

\begin{figure*}[t!]
    \centering
\includegraphics[width=0.9\linewidth]{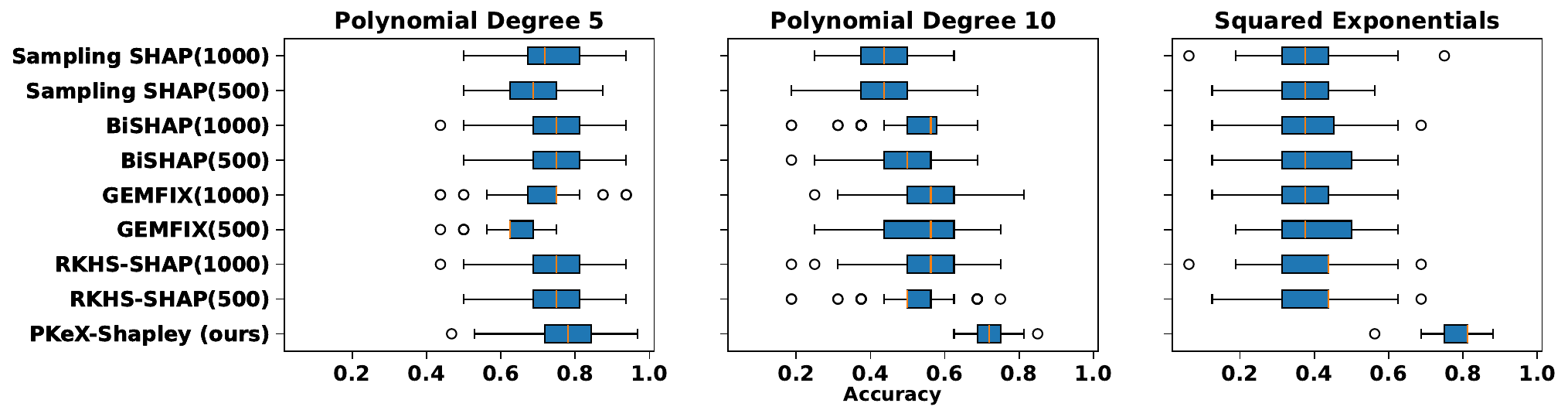}
    %\vspace{-0.6em}
    \caption{Recovery rate of true active features by each method on the three synthetic tasks.}
    % \vspace{-1em}
    \label{fig:local syn acc}
    % \vspace{-.5em}
\end{figure*}

Attribution accuracy is evaluated over 100 independent test instances by selecting the top-17 features ranked by each method and computing the fraction of truly active features recovered. Figure~\ref{fig:local syn acc} plots the average accuracy across the three tasks. PKeX-Shapley achieves competitive or superior performance in all settings. In contrast, RKHS-SHAP, GEMFIX, and Sampling SHAP exhibit decreasing accuracy as the complexity of the target function increases, especially for the degree-10 polynomial and the exponential tasks.

\begin{figure*}[t!]
    \centering
\includegraphics[width=.9\linewidth]{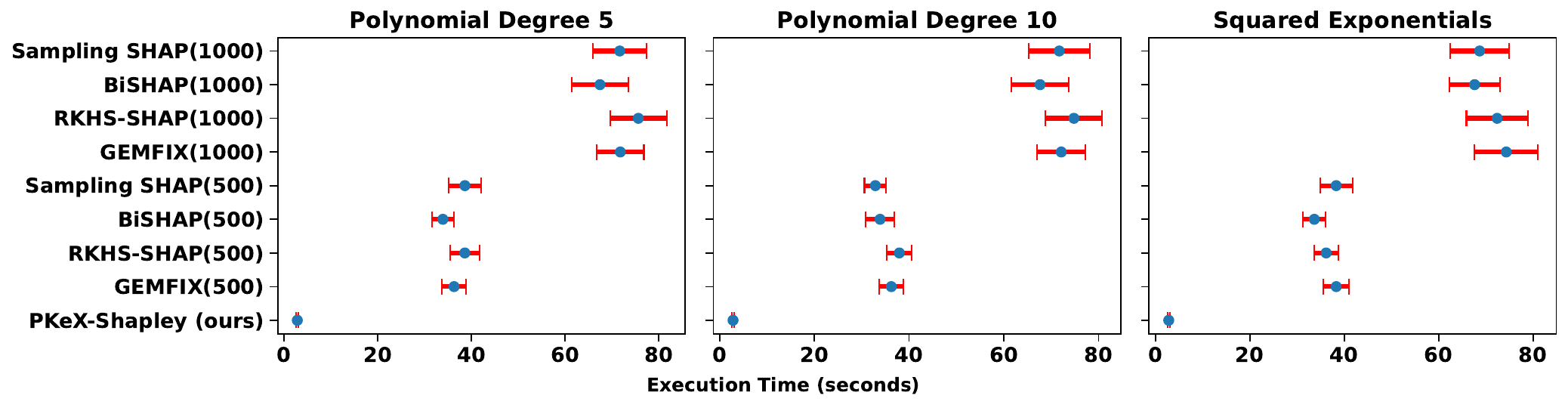}
    %\vspace{-0.6em}
    \caption{Average Per-instance explanation time for each method at $500$ and $1000$ coalition samples.}
    % \vspace{-1em}
    \label{fig:local syn time}
    \vspace{-0.8em}
\end{figure*}

Figure~\ref{fig:local syn time} depicts execution time with standard deviation across test instances. PKeX-Shapley is consistently much faster than all competing methods, even when the baselines use only 500 coalition samples. In particular, computing the full set of Shapley values with PKeX-Shapley requires, on average, slightly more than two seconds per instance, whereas the sampling-based methods require on the order of 30 seconds when limited to 500 coalition samples. This gap widens further as the number of coalition samples increases. These results highlight the substantial computational advantage of PKeX-Shapley for exact Shapley-value computation compared to sampling-based estimators.

\subsection{Explaining Distribution Discrepancy using MMD with PKeX-Shapley}
To illustrate how PKeX-Shapley can explain distributional discrepancies measured by MMD, we conduct two synthetic experiments following the standard two-sample testing setup~\citep{schrab2025unified}. Our goal is to attribute the observed MMD between two distributions to individual input variables. We present one of the synthetic experiments below, with the remaining experiments provided in Appendix~\ref{apx:mmd exp}. In all MMD experiments, we use the RBF kernel with the bandwidth selected via the median heuristic method. We generate datasets $X$ and $Z$, each comprising 20 variables. The first ten variables $X_1, \dots, X_{10}$ and $Z_1, \dots, Z_{10}$ are sampled from the same multivariate normal distribution, ensuring identical distributions. The remaining 10 variables $X_{11}, \dots, X_{20}$ differ, with variables in $X$ sampled from a multivariate normal distribution and those in $Z$ from a Student's t-distribution of the same mean. This introduces differences in higher-order moments and covariance structure, resulting in a measurable discrepancy.

\begin{figure}[t]
    \centering
  \includegraphics[width=0.7\linewidth]{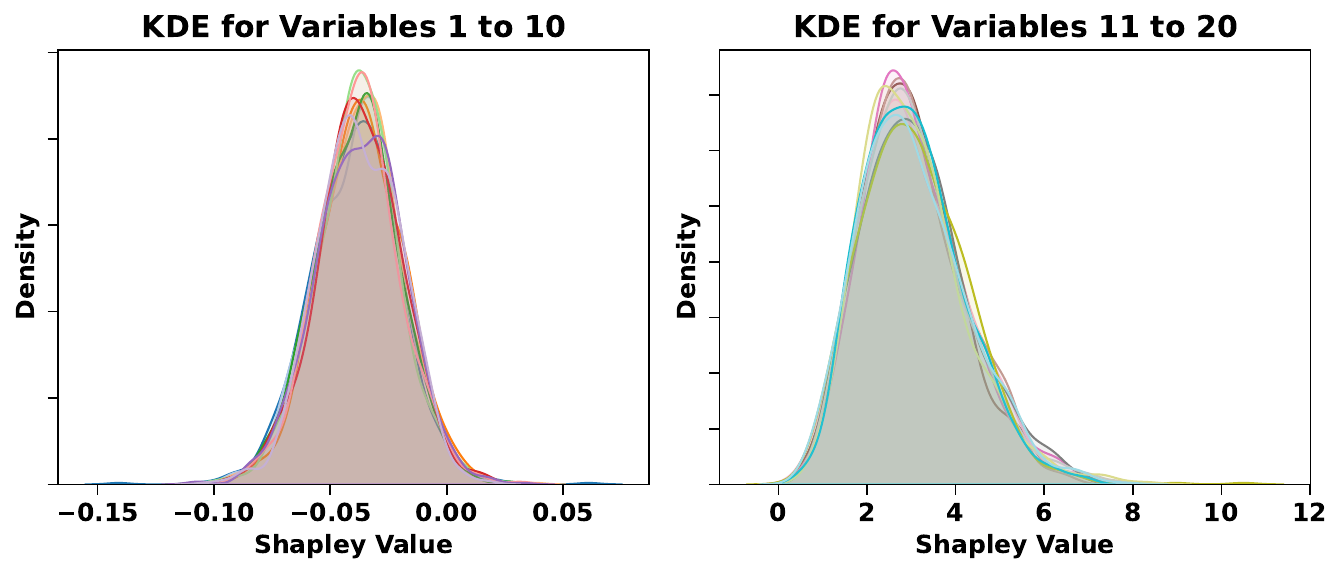}
  \caption{KDE of Shapley values. Variables $1$–$10$ have negative Shapley values (lowering MMD); $11$–$20$ are positive (raising discrepancy).}\label{fig:mmd case 1}
\end{figure}

The left subplot in Figure~\ref{fig:mmd case 1} displays the KDE plots for variables 1–10, while the right subplot corresponds to variables 11–20. For the first 10 variables, the consistently negative Shapley values indicate a reduction in the overall MMD, effectively “pulling” the discrepancy closer to a lower value. This aligns with our intuition: since these variables are identically distributed across both datasets, they should not contribute positively to the observed distributional difference. In contrast, variables 11–20 exhibit positive Shapley values, reflecting their contribution to the increase in MMD and, thus, their role in capturing the divergence between the distributions. Moreover, the KDE plots of the Shapley values within each group (variables 1–10 and 11–20, respectively) are nearly identical, which is consistent with the symmetric construction of the synthetic data. This experiment illustrates that Shapley values provide meaningful insights into the contribution of individual variables to distributional discrepancies measured by MMD.

\section{Conclusion, Limitation, and Discussion}
This work introduces PKeX-Shapley, an algorithm that computes exact Shapley values for all $d$ features of product-kernel methods in $O(d^2 n)$ time---amortised linear time per feature. The method rests on a distribution-free removal operator intrinsic to the product-kernel structure: replacing each absent feature's kernel factor with the multiplicative identity. This yields a parameter-free value function requiring no sampling, and uniquely determines the corresponding functional decomposition of the model. Recursive algorithms then compute all attributions jointly. The framework extends to kernel-based statistical discrepancies---MMD and HSIC---enabling exact and efficient interpretability of distributional differences and dependence.

The main limitation is that the algorithm applies only to product kernels. This trade-off is largely unavoidable: tractable computation requires structural constraints. We see several directions for future work. The first is to explore whether the product-kernel assumption can be relaxed without sacrificing tractability. A second is to extend the approach to higher-order attribution methods such as Shapley interaction indices~\citep{shapley_taylor, muschalik2024shapiq}. A third is to investigate whether our computational techniques can accelerate other explanation methods, including partial dependence plots and integrated gradients.

\newpage

% \newpage\clearpage
% \bibliographystyle{unsrtnat}
% \bibliography{references}

\newpage \clearpage
\appendix

%%%%%%%%%%%%%%%%%%%%%%%%%%%%%%%%%%%%%%%%%%%%%%%%%%%%%%%%%%%%

%%%%%%%%%%%%%%%%%%%%%%%%%%%%%%%%%%%%%%%%%%%%%%%%%%%%%%%%%%%%
\newpage\onecolumn
\section*{Appendix}
The appendix provides additional information and proofs related to the material presented in the main paper. It includes detailed explanations, proofs, algorithms, and experiments relevant to explaining product-kernel models. The structure of the appendix is as follows:

\begin{itemize}
    \item \textbf{Product kernels in kernel methods}: \S\ref{apx:product kernels} discusses the construction and examples of product kernels, including their definitions and properties.

    % \item \textbf{Functional ANOVA Decomposition}:\S\ref{apx:func anova} presents the functional ANOVA decomposition induced by the product kernel methods.

    \item \textbf{Newton's identities }: \S\ref{apx:newton identity} presents the main result of Newton's identities for elementary symmetric polynomials.
    
    \item \textbf{Proofs}: \S\ref{apx:proof} provides detailed proofs of key theorems and lemmas used in the main paper.
    
    % \item \textbf{Recursive and numerically stable algorithms for computing Shapley values for product-kernel learning models}: \S\ref{apx:algorithm} describes a recursive algorithm for computing Shapley values, as well as an adjusted algorithm for numerical stability and efficiency.

    % \item \textbf{Additivity of explanations for learning models, MMD and HSIC}: \S\ref{apx:additivity} discusses the additivity of explanations for the product-kernel methods studied in this paper.
	
    % \item \textbf{Shapley value attribution for HSIC with two multivariate variables} \S \ref{apx:hsic two multivariate} discusses the Shapley value attribution for HSIC when both random variables are multivariate.

    \item \textbf{Interpolation-Based ESP Recovery Methods}: 
    Section~\ref{apx:interpolation} investigates whether interpolation-based approaches, similar to those used in TreeSHAP, can be applied to product-kernel methods and examines the numerical stability of the resulting algorithms.

    \item \textbf{Experiments}: \S\ref{apx:experiments} reports experimental setup and extra experiments on explaining product-kernel models.

    \item \textbf{FAQ}: \S\ref{apx:faq} responds to main question about PKeX-Shapley.

\end{itemize}

%%%%%%%%%%%%%%%%%%%%%%%%%%%%%%%%%%%%%%%%%%%%%%%%%%%%%%%%%%%%
%%%%%%%%%%%% Product Kernels
%%%%%%%%%%%%%%%%%%%%%%%%%%%%%%%%%%%%%%%%%%%%%%%%%%%%%%%%%%%%
\section{Product Kernels in Kernel Methods}\label{apx:product kernels}
Product kernels provide a powerful mechanism for constructing high-dimensional similarity measures by combining kernels defined on individual dimensions or feature subsets. This section discusses key examples of product kernels. 

\paragraph{Radial Basis Function (RBF) Kernels as Product Kernels}  

The RBF kernel is a canonical example of product kernels. The RBF kernel is defined based on a distance metric between two instances, with the two well-known metrics being Euclidean (L2 norm) and Manhattan distances (L1 norm). We refer to the former as RBF, and the latter as Laplacian RBF to distinguish these two kernel functions. In addition, when we have only one kernel bandwidth parameter $\sigma$, the RBF kernel is referred to as \textit{isotropic}. The RBF kernel with both distance metrics inherently decomposes into products of univariate kernels across dimensions:

\begin{itemize}  
    \item \textbf{RBF Kernel}:  
    \begin{align*}  
    K_{\text{RBF}}(\x, \z) = \exp\left(-\frac{\|\x - \z\|^2}{2\sigma^2}\right) 
    = \prod_{i=1}^d \exp\left(-\frac{(x_i - z_i)^2}{2\sigma^2}\right).  
    \end{align*}  
    \item \textbf{Laplacian RBF Kernel}:  
    \[  
    K_{\text{Laplacian RBF}}(\x, \z) = \exp\left(-\frac{\|\x - \z\|_1}{\sigma}\right) = \prod_{i=1}^d \exp\left(-\frac{|x_i - z_i|}{\sigma}\right).  
    \]  
\end{itemize}  

When alternative distance metrics are incorporated into the RBF kernel, such as the Mahalanobis distance, which involves a covariance matrix, the resulting kernel might lose its product decomposition.

\paragraph{Automatic Relevance Determination (ARD) in Gaussian Processes}  
ARD extends RBF kernels by assigning independent length-scale parameters \(\sigma_i\) to each dimension:  
\[  
K_{\text{ARD}}(\x, \z) = \exp\left(-\sum_{i=1}^d \frac{(x_i - z_i)^2}{2\sigma_i^2}\right) = \prod_{i=1}^d \exp\left(-\frac{(x_i - z_i)^2}{2\sigma_i^2}\right).  
\]  

The ARD is extensively used in Gaussian processes for feature selection via learned \(\sigma_i\), and to enhance interpretability and adaptability. ARD is also referred to as \textit{anisotropic}.  

\paragraph{Cauchy Kernel}  
The Cauchy kernel, inspired by the Cauchy distribution, is another example of a product kernel:  
\[  
K_{\text{Cauchy}}(\x, \z) = \prod_{i=1}^d \frac{1}{1 + \frac{(x_i - z_i)^2}{\sigma^2}}.  
\]

\paragraph{Product of Base Kernels}  
A popular way for constructing product kernels is by first defining a base kernel for each individual feature and then computing the overall kernel function by multiplying the base kernels over individual features. The product of PSD kernels remains PSD by the Schur product theorem:  
\[  
K(\x, \z) = \prod_{i=1}^d K_i(x_i, z_i).  
\]  

This type of kernel introduces flexibility in combining base kernels while maintaining validity as a (product) kernel function.

\section{Newton's Identities}\label{apx:newton identity}

To explain Newton's identities, we begin with a specific set of variables $\cZ_4 = \{z_1, z_2, z_3, z_4\}$ before generalizing it to sets of arbitrary size $\cZ_d = \{z_1, z_2, \ldots, z_d\}$.
The \textit{elementary symmetric polynomials} (ESPs) of degree $q$ is defined as

\[
e_q(\cZ_4) = \sum_{1 \leq i_1 < i_2 < \cdots < i_q \leq 4} z_{i_1} z_{i_2} \cdots z_{i_q},
\]

with the conventions $e_0(\cZ_4) = 1$ and $e_q(\cZ_4) = 0$ for $q > 4$. For example:
\begin{align*}
&e_1(\cZ_4) = z_1 + z_2 + z_3 + z_4, \cr
&e_2(\cZ_4) = z_1z_2 + z_1z_3 + z_1z_4 + z_2z_3 + z_2z_4 + z_3z_4, \cr
&e_4(\cZ_4) = z_1z_2z_3z_4.
\end{align*}

The \textit{power sum} of degree $r$ is given by
\[
p_r(\cZ_4) = z_1^r + z_2^r + z_3^r + z_4^r.
\]
In particular, $p_1(\cZ_4) = e_1(\cZ_4)$ and, for example, $p_2(\cZ_4) = z_1^2 + z_2^2 + z_3^2 + z_4^2$.
Then, Newton's identities relate the ESPs to the power sum recursively. For $q \geq 1$,
\[
e_q(\cZ_4) = \frac{1}{q} \sum_{r=1}^{q} (-1)^{r-1} e_{q-r}(\cZ_4)\, p_r(\cZ_4).
\]
For the set $\cZ_4 = \{z_1, z_2, z_3, z_4\}$, the identities yield:
{\small 
\begin{align*}
e_1(\cZ_4) &= \frac{1}{1}[ e_0(\cZ_4) \, p_1(\cZ_4) ] = p_1(\cZ_4) = z_1 + z_2 + z_3 + z_4, \\
e_2(\cZ_4) &= \frac{1}{2}\Big[ e_1(\cZ_4) \, p_1(\cZ_4) - e_0(\cZ_4) \, p_2(\cZ_4) \Big] \\
&\quad = \frac{(z_1 + z_2 + z_3 + z_4)^2 - (z_1^2 + z_2^2 + z_3^2 + z_4^2)}{2}, \\
e_3(\cZ_4) &= \frac{1}{3}\Big[ e_2(\cZ_4) \, p_1(\cZ_4) - e_1(\cZ_4) \, p_2(\cZ_4) + e_0(\cZ_4) \, p_3(\cZ_4) \Big], \\
e_4(\cZ_4) &= \frac{1}{4}\Big[ e_3(\cZ_4) \, p_1(\cZ_4) - e_2(\cZ_4) \, p_2(\cZ_4) \\ 
& \qquad + e_1(\cZ_4) \, p_3(\cZ_4) - e_0(\cZ_4) \, p_4(\cZ_4) \Big].
\end{align*}
}
The identities presented above can be extended to sets of arbitrary size $\cZ_d = \{z_1, z_2, \dots, z_d\}$ as
\[
e_q(\cZ_d) = \frac{1}{q} \sum_{r=1}^{q} (-1)^{r-1} e_{q-r}(\cZ_d)\, p_r(\cZ_d), \quad \text{for } q \geq 1,
\]
with $e_0(\cZ_d) = 1$ and $e_q(\cZ_d) = 0$ if $q > d$ or $q < 0$.

%%%%%%%%%%%%%%%%%%%%%%%%%%%%%%%%%%%%%%%%%%%%%%%%%%%%%%%%%%%%
%%%%%%%%%%%% Proofs
%%%%%%%%%%%%%%%%%%%%%%%%%%%%%%%%%%%%%%%%%%%%%%%%%%%%%%%%%%%%
\section{Proofs}\label{apx:proof}
This section provides the detailed proofs of the theoretical results presented in the main paper. First of all, we present a theorem that plays a key role in the other proofs. 

\begin{Proposition}\label{th:multiplying kernels}

Let the kernel function for feature $j$ be denoted by $k_j(x_j,x'_j)$, and $\prod_{j \in \emptyset} (k_j(x_j,x'_j)-1) = 1$ by convention. Then, the following equation holds for product kernel functions:
\[
\prod_{j \in \cD} k_j(x_j, x'_j) 
= \sum_{\cS \subseteq \cD} \prod_{j \in \cS}(k_j(x_j, x'_j)-1).
\]
\end{Proposition}

\paragraph{Proof.}
We prove the statement by induction on $|\cD|$.

\textbf{Base case $(|\cD|=1)$:} Let $\cD=\{j\}$. Then
$
k_j(x_j,x'_j)=1+(k_j(x_j,x'_j)-1),
$
as desired.

\textbf{Inductive step:} Assume the equation holds for any set of size $n$. Consider a set $\cD$ with size $n+1$, and let $a \in \cD$. We write:
\[
\prod_{j \in \cD} k_j(x_j,x'_j)
= k_a(x_a,x'_a)\prod_{j \in \cD\setminus\{a\}}k_j(x_j,x'_j).
\]
Using the induction hypothesis for $\cD\setminus\{a\}$, we have
\[
\prod_{j \in \cD} k_j(x_j,x'_j) = k_a(x_a,x'_a)\sum_{\cT\subseteq \cD\setminus\{a\}}\prod_{j\in \cT}(k_j(x_j,x'_j)-1).
\]

Since $k_a(x_a,x'_a)=(k_a(x_a,x'_a)-1)+1$, expanding gives:
\[
\prod_{j \in \cD} k_j(x_j,x'_j) = \left[(k_a(x_a,x'_a)-1)+1\right]\sum_{\cT\subseteq \cD\setminus\{a\}}\prod_{j\in \cT}(k_j(x_j,x'_j)-1).
\]

Expanding this product, we obtain:
\[
\prod_{j \in \cD} k_j(x_j,x'_j) = \sum_{\cT\subseteq \cD\setminus\{a\}}(k_a(x_a,x'_a)-1)\prod_{j\in \cT}(k_j(x_j,x'_j)-1)
\: + \sum_{\cT\subseteq \cD\setminus\{a\}}\prod_{j\in \cT}(k_j(x_j,x'_j)-1).
\]

The first sum on the right-hand side of the equation covers all subsets that contain $a$, and the second sum covers all subsets that do not contain $a$. Thus, together they sum over all subsets $\mathcal{\cS} \subseteq \mathcal{\cD}$:
\[
\prod_{j \in \cD} k_j(x_j,x'_j)= \sum_{\cS\subseteq\cD}\prod_{j\in \cS}(k_j(x_j,x'_j)-1).
\]

Thus, by induction, the theorem is proved. \qed

We now provide the proofs for all the theorems and lemmas in the manuscript.

%%%%%%%%%%%%%%%%%%%%%%%%%%
\subsection*{Proof of Proposition~\ref{prop:func-decomp}}

Using the product kernel, the decision function can be rewritten as:
\[
f(\x) = \sum_{i=1}^{n} \alpha_i \prod_{j \in \cD} k_j(x_j,x^{(i)}_j).
\]
Expanding the product using Proposition \ref{th:multiplying kernels}:
\[
\prod_{j \in \cD} k_j(x_j,x^{(i)}_j) = \sum_{\cS \subseteq \cD} \prod_{j \in \cS} \big(k_j(x_j,x^{(i)}_j) - 1\big).
\]
Substituting this expansion into the decision function:
\[
f(\x) = \sum_{i=1}^{n} \alpha_i \sum_{\cS \subseteq \cD} \prod_{j \in \cS} \big( k_j(x_j,x^{(i)}_j) - 1\big).
\]
Rearranging the summations:
\[
f(\x) = \sum_{\cS \subseteq \cD} \sum_{i=1}^{n} \alpha_i \prod_{j \in \cS} \big( k_j(x_j,x^{(i)}_j) - 1\big).
\]
This establishes the functional decomposition:
\[
f(\x) = \sum_{\cS \subseteq \cD} f_\cS(\x_{\cS}),
\]
where each term is defined as:
\[
f_\cS(\x_{\cS}) = \sum_{i=1}^{n} \alpha_i \prod_{j \in \cS} \big( k_j(x_j,x^{(i)}_j) - 1 \big).
\]
This establishes the existence of the decomposition. To see uniqueness relative to the chosen removal family, let $R_\cS f := P_{\cD\setminus\cS}f$. By the canonical additive decomposition (CAD) construction of \citet{gevaert2024unifying}, the induced components satisfy
\[
R_\cS f(\x) = \sum_{\cT\subseteq\cS} f_\cT(\x_\cT), \qquad \cS\subseteq\cD.
\]
By M\"{o}bius inversion on the subset lattice,
\[
f_\cS(\x_\cS) = \sum_{\cT\subseteq\cS}(-1)^{|\cS|-|\cT|} R_\cT f(\x).
\]
For the product-kernel removal, $R_\cT f(\x) = \alphab^\top k_\cT(\X_\cT,\x_\cT)$, so
\[
f_\cS(\x_\cS)
= \alphab^\top \sum_{\cT\subseteq\cS}(-1)^{|\cS|-|\cT|} \bigodot_{j\in\cT}\z_j
= \sum_{i=1}^n \alpha_i \prod_{j\in\cS}\bigl(k_j(x_j,x_j^{(i)})-1\bigr),
\]
where the last equality is the standard inclusion-exclusion identity $\sum_{\cT\subseteq\cS}(-1)^{|\cS|-|\cT|}\prod_{j\in\cT}z_j = \prod_{j\in\cS}(z_j-1)$. Thus these are the unique CAD components induced by the chosen removal family. \qed

%%%%%%%%%%%%%%%%%%%%%%%%%%%%%%%%
\subsection*{Proof of Proposition~\ref{prop:func-decomp-value}}
By Proposition~\ref{prop:func-decomp}, the product-kernel predictor admits the functional decomposition
\[
f(\x)=\sum_{\cT\subseteq\cD} f_\cT(\x_\cT),
\qquad
f_\cT(\x_\cT)
=
\sum_{i=1}^n \alpha_i
\prod_{j\in \cT}\big(k_j(x_j,x_j^{(i)})-1\big).
\]
Following the decomposition-based removal view of \citet{gevaert2024unifying}, removing the variables in
\(\cD\setminus \cS\) retains only those decomposition components whose support is contained in \(\cS\). Hence,
\[
P_{\cD\setminus \cS}(f)(\x)
:=
\sum_{\cT\subseteq \cS} f_\cT(\x_\cT).
\]
Substituting the product-kernel components gives
\begin{align*}
P_{\cD\setminus \cS}(f)(\x)
&=
\sum_{\cT\subseteq \cS}
\sum_{i=1}^n \alpha_i
\prod_{j\in \cT}\big(k_j(x_j,x_j^{(i)})-1\big) \\
&=
\sum_{i=1}^n \alpha_i
\sum_{\cT\subseteq \cS}
\prod_{j\in \cT}\big(k_j(x_j,x_j^{(i)})-1\big).
\end{align*}
For each fixed \(i\), the inner sum expands as
\[
\sum_{\cT\subseteq \cS}
\prod_{j\in \cT}\big(k_j(x_j,x_j^{(i)})-1\big)
=
\prod_{j\in \cS}
\left(1+\big(k_j(x_j,x_j^{(i)})-1\big)\right)
=
\prod_{j\in \cS} k_j(x_j,x_j^{(i)}).
\]
Therefore,
\[
P_{\cD\setminus \cS}(f)(\x)
=
\sum_{i=1}^n \alpha_i
\prod_{j\in \cS} k_j(x_j,x_j^{(i)})
=
\alphab^\top k_\cS(\X_\cS,\x_\cS).
\]
Thus, for product-kernel models, decomposition-based removal is algebraically equivalent to replacing all absent kernel factors by the multiplicative identity \(1\). This is natural since $k_j = 1$ is the neutral element for the multiplicative structure.

Following the pointwise cooperative-game construction induced by the removal operator~\cite[Definition 40]{gevaert2024unifying}, we define
\[
v_{\x}(\cS)
=
P_{\cD\setminus \cS}(f)(\x)-P_{\cD}(f)(\x).
\]
Since
\[
P_{\cD}(f)(\x)=f_\emptyset=\sum_{i=1}^n\alpha_i,
\]
we obtain
\[
v_{\x}(\cS)
=
\alphab^\top k_\cS(\X_\cS,\x_\cS)-f_\emptyset.
\]
In particular,
\[
v_{\x}(\emptyset)=f_\emptyset-f_\emptyset=0,
\qquad
v_{\x}(\cD)=f(\x)-f_\emptyset.
\]
This proves the claimed value function.

% By Proposition~\ref{prop:func-decomp} and \citet[Theorem~19]{gevaert2024unifying}, the functional decomposition induces a unique removal operator $P_\cT$ that sets $k_j=1$ for each $j\in\cT$, giving $P_\cT(f)(\x) = \alphab^\top k_{\cD\setminus\cT}(\X_{\cD\setminus\cT},\x_{\cD\setminus\cT})$. The value function is defined as
% \[
% v_{\x}(\cS) \;:=\; P_{\cD\setminus\cS}(f)(\x) - P_{\cD}(f)(\x),
% \]
% where the baseline $P_{\cD}(f)(\x) = f_\emptyset = \sum_{i=1}^n\alpha_i$ is subtracted to ensure $v_{\x}(\emptyset)=0$. It remains to show that $P_{\cD\setminus\cS}(f)(\x) = \alphab^\top k_\cS(\X_\cS,\x_\cS)$.

% Starting from the functional decomposition in Proposition~\ref{prop:func-decomp}:
% \begin{align*}
% P_{\cD\setminus\cS}(f)(\x) = \sum_{\cT\subseteq\cS} f_\cT(\x_\cT)
% &= \sum_{i=1}^n \alpha_i \sum_{\cT\subseteq\cS}\prod_{j\in\cT}\big(k_j(x_j,x^{(i)}_j)-1\big).
% \end{align*}
% For fixed $i$, the inner sum enumerates, for each $j\in\cS$, the binary choice between $1$ (when $j\notin\cT$) and $k_j(x_j,x^{(i)}_j)-1$ (when $j\in\cT$). Summing over all $2^{|\cS|}$ such choices via the binomial identity gives $\prod_{j\in\cS} k_j(x_j,x^{(i)}_j)$, so $P_{\cD\setminus\cS}(f)(\x) = \alphab^\top k_\cS(\X_\cS,\x_\cS)$.

% Therefore $v_{\x}(\cS) = \alphab^\top k_\cS(\X_\cS,\x_\cS) - f_\emptyset$, which satisfies $v_{\x}(\emptyset) = \alphab^\top k_\emptyset - f_\emptyset = f_\emptyset - f_\emptyset = 0$ and $v_{\x}(\cD) = \alphab^\top k_\cD(\X,\x) - f_\emptyset = f(\x) - f_\emptyset$. \qed

%%%%%%%%%%%%%%%%%%%%%%%%%%%%%%%%
\subsection*{Proof of \Cref{th:shapley recursive}}
We can write the Shapley value as follows:
\begin{align*}
\phi_j^{\x}
&:= \sum_{\cS \subseteq \cD \setminus \{j\}}
\mu(|\cS|)
\big(
v_{\x}(\cS\cup \{j\}) - v_{\x}(\cS)
\big) \\
&= \sum_{q=0}^{d-1} \mu(q)
\sum_{\substack{\cS \subseteq \cD \setminus \{j\} \\ |\cS| = q}}
\big(
v_{\x}(\cS\cup \{j\}) - v_{\x}(\cS)
\big).
\end{align*}
We substitute the functional decomposition value function $v_{\x}(\cS) = \alphab^\top k_\cS(\X_\cS,\x_\cS)$. The marginal contribution of feature $j$ to coalition $\cS$ is:
\begin{align*}
v_{\x}(\cS\cup\{j\}) - v_{\x}(\cS)
&= \alphab^\top\big(k_{\cS\cup\{j\}}(\X_{\cS\cup\{j\}},\x_{\cS\cup\{j\}}) - k_\cS(\X_\cS,\x_\cS)\big) \\
&= \alphab^\top\big((k_j(\X_j,x_j)-\onevec)\odot k_\cS(\X_\cS,\x_\cS)\big),
\end{align*}
where we used the product kernel factorization $k_{\cS\cup\{j\}} = k_j \cdot k_\cS$ and factored out $k_\cS$.

Let $\z_i := k_i(\X_i, x_i)$ and $\cZ=\{\z_1,\ldots,\z_d\}$. Substituting into the Shapley sum:
\begin{align}\label{eq:sv kernel func decomp first}
\phi_j^{\x}
&=
\alphab^\top \Bigg(
\big(k_j(\X_j,x_j) - \onevec\big)
\odot
\sum_{q=0}^{d-1} \mu(q)
\sum_{\substack{\cS \subseteq \cD \setminus \{j\} \\ |\cS| = q}}
k_\cS(\X_\cS,\x_\cS)
\Bigg),
\end{align}
where all products and subtractions are element-wise. By definition of the elementary symmetric polynomials (ESPs) over the vectors $\{\z_i\}$:
\[
\sum_{\substack{\cS \subseteq \cD \setminus \{j\} \\ |\cS| = q}} k_\cS(\X_\cS,\x_\cS)
= \sum_{\substack{\cS \subseteq \cD \setminus \{j\} \\ |\cS| = q}} \bigodot_{i\in\cS}\z_i = e_q(\cZ_{-j}).
\]
Substituting back into~\eqref{eq:sv kernel func decomp first}:
\begin{align*}
\phi_j^{\x}
&=
\alphab^\top \Bigg(
\big( \z_j - \onevec \big)
\odot
\sum_{q=0}^{d-1} \mu(q)\, e_q(\cZ_{-j})
\Bigg).
\end{align*}
The ESP vectors $e_q(\cZ_{-j})$ can be recursively computed via Newton's identities, with base case $e_0(\cZ_{-j}) = \onevec$ (the all-ones vector in $\mathbb{R}^n$) and for $q \ge 1$:
\begin{align*}
e_q(\cZ_{-j})
&=
\frac{1}{q}
\sum_{r=1}^{q}
(-1)^{r-1}
e_{q-r}(\cZ_{-j})
\odot
p_r(\cZ_{-j}),
\end{align*}
where $p_r(\cZ) = \sum_{\z_i\in\cZ}\z_i^r$ and the power is applied element-wise. This completes the proof.

%%%%%%%%%%%%%%%%%%%%%%%%%%%%%%%%
\subsection*{Proof of Proposition~\ref{prop:omega_prefix_suffix}}
To justify the decomposition of $e_q(\cZ_{-j})$ into prefix and suffix contributions, we start from the polynomial representation introduced in the main body. Consider the leave-one-out polynomial
\[
p_{-j}(s)
:=
\bigodot_{i\in \cD\setminus\{j\}} \big(\onevec + \z_i\, s\big),
\]
whose coefficient of $s^q$ is, by definition, the degree-$q$ ESP $e_q(\cZ_{-j})$.
Since $\cD\setminus\{j\} = \{1,\dots,j-1\} \cup \{j+1,\dots,d\}$ is a disjoint union, the polynomial $p_{-j}(s)$ factorizes as the product of a prefix and a suffix polynomial,
\[
p_{-j}(s)
=
\left(
\bigodot_{i=1}^{j-1} \big(\onevec + \z_i\, s\big)
\right)\bigodot
\left(
\bigodot_{i=j+1}^{d} \big(\onevec + \z_i\, s\big)
\right).
\]
Using the definitions of the prefix and suffix polynomials, this can be written as
\[
p_{-j}(s)
=
\left(
\sum_{t=0}^{j-1} \bGamma_{j-1,t}\, s^t
\right) \odot
\left(
\sum_{u=0}^{d-j} \bR_{j+1,u}\, s^u
\right).
\]
Expanding the product yields
\[
p_{-j}(s)
=
\sum_{t=0}^{j-1}\sum_{u=0}^{d-j}
\big(\bGamma_{j-1,t}\odot \bR_{j+1,u}\big)\, s^{t+u}.
\]

Collecting terms with the same total degree $q=t+u$, the coefficient of $s^q$ is therefore given by
\[
e_q(\cZ_{-j})
=
\sum_{\substack{t,u\ge 0\\ t+u=q}}
\bGamma_{j-1,t}\odot \bR_{j+1,u},
\]
where only indices $t\in\{0,\dots,j-1\}$ and $u\in\{0,\dots,d-j\}$ contribute nonzero terms. This establishes the claimed prefix--suffix decomposition of the ESPs over $\cZ_{-j}$.

Substituting this expression into $\sum_{q=0}^{d-1} \mu(q)\, e_q(\cZ_{-j})$ yields
\begin{align*}
\sum_{q=0}^{d-1} \mu(q)\, e_q(\cZ_{-j}) &=
\sum_{q=0}^{d-1}\mu(q)
\sum_{\substack{t,u\ge 0\\ t+u=q}}
\bGamma_{j-1,t} \odot \bR_{j+1,u} \\
&=
\sum_{t=0}^{j-1}\sum_{u=0}^{d-j}
\mu(t+u)\,
\big(\bGamma_{j-1,t} \odot \bR_{j+1,u}\big).    
\end{align*}

For fixed $t$, the factor $\bGamma_{j-1,t}$ does not depend on $u$ and can be factored out of the inner sum, giving
\[
\sum_{q=0}^{d-1} \mu(q)\, e_q(\cZ_{-j}) =
\sum_{t=0}^{j-1}
\bGamma_{j-1,t} \odot
\left(
\sum_{u=0}^{d-j} \mu(t+u)\, \bR_{j+1,u}
\right).
\]
Recognizing the inner term as $\bG_{j+1,t}$ completes the proof.

%%%%%%%%%%%%%%%%%%%%%%
%%%%%% Proof of Proposition 4
%%%%%%%%%%%%%%%%%%%%%%%
\subsection*{Proof of Proposition~\ref{prop:time_memory_complexity}}
The prefix coefficients $\bGamma_{j,t}$ arise as the polynomial coefficients of the prefix polynomial
\[
p^{\mathrm{pref}}_j(s)
=
\bigodot_{i=1}^{j} (\onevec+\z_i s)
=
\sum_{t=0}^{j} \bGamma_{j,t}s^t .
\]
Multiplying $p^{\mathrm{pref}}_{j-1}(s)$ by $(\onevec+\z_j s)$ and matching coefficients of equal powers of $s$ yields the recurrence
\[
\bGamma_{j,t}
=
\bGamma_{j-1,t}
+
\z_j\odot \bGamma_{j-1,t-1},
\qquad j=1,\dots,d,
\]
with the convention $\bGamma_{j,-1}=\mathbf{0}$. This allows all prefix coefficients to be computed in a single forward pass. The aggregated suffix quantities are defined from the suffix ESPs by
\[
\bG_{j,t}
:=
\sum_{u=0}^{d-j+1}\mu(t+u)\,\bR_{j,u}.
\]
The suffix ESP $\bR$ has a similar recurrence to $\bGamma$. Substituting the suffix recurrence
$\bR_{j,u}=\bR_{j+1,u}+\z_j\odot \bR_{j+1,u-1}$
into this definition gives
\[
\begin{aligned}
\bG_{j,t}
&=
\sum_{u}\mu(t+u)\Big(\bR_{j+1,u}+\z_j\odot \bR_{j+1,u-1}\Big) \\
&=
\underbrace{\sum_{u}\mu(t+u)\bR_{j+1,u}}_{=\,\bG_{j+1,t}}
+
\z_j\odot
\underbrace{\sum_{u\ge 1}\mu(t+u)\bR_{j+1,u-1}}_{=\,\bG_{j+1,t+1}},
\end{aligned}
\]
where the second identity follows from the index shift $u\mapsto u-1$. Consequently, $\bG$ satisfies the backward recursion
\[
\bG_{j,t}
=
\bG_{j+1,t}
+
\z_j\odot \bG_{j+1,t+1},
\qquad j=d,\dots,1,
\]
with boundary condition $\bG_{d+1,t}=\mu(t)\onevec$ induced by the empty-suffix polynomial.

We set $\mu(q)=0$ for $q\notin\{0,\ldots,d-1\}$ so the boundary terms in the backward recurrence vanish automatically. Using these recursive formulations, computing and storing the triangular tables $\bGamma$ and $\bG$ requires $O(d^2 n)$ time and $O(d^2 n)$ memory, since each table entry $\bGamma_{j,t}$ or $\bG_{j,t}$ is a vector in $\mathbb{R}^n$. Given $\bGamma$ and $\bG$, each Shapley value $\phi_j^{\x}$ can be computed on the fly in $O(dn)$ time and $O(n)$ additional memory. Therefore, the overall time complexity for computing all $d$ Shapley values is $O(d^2 n)$.

%%%%%%%%%%%%%%%%%%%%%%%%%%%%%%%%
\subsection*{Proof of \Cref{th:mmd func decomp}}
\begin{itemize}[leftmargin=8pt]
    \item[(ii)] We begin by expressing the full kernel function as a product of feature-wise kernels:
    \begin{equation*}
        k(\x^{(i)}, \x^{(j)}) = \prod_{q \in \cD} k_q(x^{(i)}_q, x^{(j)}_q).
    \end{equation*}
    Using Proposition \ref{th:multiplying kernels}, we expand the product:
    \begin{equation*}
        \prod_{q \in \cD} k_q(x^{(i)}_q, x^{(j)}_q) = \sum_{\cS \subseteq \cD} \prod_{q \in \cS} \big( k_q(x_q^{(i)}, x_q^{(j)}) - 1 \big).
    \end{equation*}
    Substituting this into the MMD formulation for \( k(\x^{(i)}, \x^{(j)}) \), \( k(\z^{(i)}, \z^{(j)}) \), and \( k(\x^{(i)}, \z^{(j)}) \) yields
    \begin{align*}
        &\widehat{\text{MMD}}^2(\mathbb{P}, \mathbb{Q}) = \frac{1}{n(n-1)} \sum_{i \neq j} \sum_{\cS \subseteq \cD} \prod_{q \in \cS} \big( k_q(x^{(i)}_q, x^{(j)}_q) - 1 \big) \nonumber \\
        &\quad + \frac{1}{m(m-1)} \sum_{i \neq j} \sum_{\cS \subseteq \cD} \prod_{q \in \cS} \big( k_q(z^{(i)}_q, z^{(j)}_q) - 1 \big) - \frac{2}{nm} \sum_{i, j} \sum_{\cS \subseteq \cD} \prod_{q \in \cS} \big( k_q(x^{(i)}_q, z^{(j)}_q) - 1 \big).
    \end{align*}
    By the linearity of summation, we have
    \begin{align*}
        \widehat{\text{MMD}}^2(\mathbb{P}, \mathbb{Q}) &= \sum_{\cS \subseteq \cD} \bigg[ \frac{1}{n(n-1)} \sum_{i \neq j} \prod_{q \in \cS} \big( k_q(x^{(i)}_q, x^{(j)}_q) - 1 \big) \nonumber \\
        &\quad + \frac{1}{m(m-1)} \sum_{i \neq j} \prod_{q \in \cS} \big( k_q(z^{(i)}_q, z^{(j)}_q) - 1 \big)  - \frac{2}{nm} \sum_{i, j} \prod_{q \in \cS} \big( k_q(x^{(i)}_q, z^{(j)}_q) - 1 \big) \bigg].
    \end{align*}
    This establishes the functional decomposition of MMD:
    \begin{equation*}
        \widehat{\text{MMD}}^2(\mathbb{P}, \mathbb{Q}) = \sum_{\cS \subseteq \cD} f^{\text{MMD}}_{\cS}(\mathbb{P}, \mathbb{Q}),
    \end{equation*}
    where \( f^{\text{MMD}}_{\cS} \) is defined as
    \begin{align*}
        f^{\text{MMD}}_{\cS}(\mathbb{P}, \mathbb{Q}) &= \frac{1}{n(n-1)} \sum_{i \neq j} \prod_{q \in \cS} \big( k_q(x^{(i)}_q, x^{(j)}_q) - 1 \big)
        \cr & + \frac{1}{m(m-1)} \sum_{i \neq j} \prod_{q \in \cS} \big( k_q(z^{(i)}_q, z^{(j)}_q) - 1 \big)  - \frac{2}{nm} \sum_{i, j} \prod_{q \in \cS} \big( k_q(x^{(i)}_q, z^{(j)}_q) - 1 \big).
    \end{align*}

    \item[(i)] Following the same reasoning as in Proposition~\ref{prop:func-decomp-value}, the functional decomposition of MMD in part~(ii) induces a removal operator that sets $k_j=1$ for features in the absent set. The corresponding value function is defined as $v_{\text{MMD}}(\cS) := P_{\cD\setminus\cS}(\widehat{\text{MMD}}^2) - P_{\cD}(\widehat{\text{MMD}}^2)$. Since $P_{\cD}(\widehat{\text{MMD}}^2) = f^{\text{MMD}}_\emptyset = 0$ (setting all kernels to~$1$ yields equal cross-terms that cancel), this reduces to
    \begin{equation*}
        v_{\text{MMD}}(\cS) = \sum_{\cT \subseteq \cS} f^{\text{MMD}}_{\cT}(\mathbb{P}, \mathbb{Q}).
    \end{equation*}

    Applying this to MMD, we define:
    \begin{align*}
        v_{\text{MMD}}(\cS) &= \sum_{\cT \subseteq \cS} \frac{1}{n(n-1)} \sum_{i \neq j} \prod_{q \in \cT} \big( k_q(x^{(i)}_q, x^{(j)}_q) - 1 \big) \cr
        &\qquad + \frac{1}{m(m-1)} \sum_{i \neq j} \prod_{q \in \cT} \big( k_q(z^{(i)}_q, z^{(j)}_q) - 1 \big)  - \frac{2}{nm} \sum_{i, j} \prod_{q \in \cT} \big( k_q(x^{(i)}_q, z^{(j)}_q) - 1 \big) \cr 
        & = \frac{1}{n(n-1)} \sum_{i \neq j} \sum_{\cT \subseteq \cS} \prod_{q \in \cT} \big( k_q(x^{(i)}_q, x^{(j)}_q) - 1 \big) \cr 
        & + \frac{1}{m(m-1)} \sum_{i \neq j} \sum_{\cT \subseteq \cS} \prod_{q \in \cT} \big( k_q(z^{(i)}_q, z^{(j)}_q) - 1 \big)  - \frac{2}{nm} \sum_{i, j} \sum_{\cT \subseteq \cS} \prod_{q \in \cT} \big( k_q(x^{(i)}_q, z^{(j)}_q) - 1 \big).
    \end{align*}

    By applying Proposition~\ref{th:multiplying kernels} to the inner sum, we can simplify the above equation as:
    \begin{equation*}
        v_{\text{MMD}}(\cS) = \frac{1}{n(n-1)} \sum_{i \neq j} k_{\cS}(\x^{(i)}_{\cS}, \x^{(j)}_{\cS}) + \frac{1}{m(m-1)} \sum_{i \neq j} k_{\cS}(\z^{(i)}_{\cS}, \z^{(j)}_{\cS}) - \frac{2}{nm} \sum_{i, j} k_{\cS}(\x^{(i)}_{\cS}, \z^{(j)}_{\cS}),
    \end{equation*}
    which completes the proof. \qed
\end{itemize}

%%%%%%%%%%%%%%%%%%%%%%%%%%%%%%%%
\subsection*{Proof of Proposition~\ref{prop:mmd sv}}
By substituting \( v_{\text{MMD}}(\cS) \) into the Shapley value formula, we obtain:
\begin{align*}
    \phi_q^{\text{MMD}} &= \sum_{r=0}^{d-1} \mu(r) \sum_{\substack{\cS \subseteq \cD \setminus \{q\} \\ |\cS| = r}} \bigg( v_{\text{MMD}}(\cS \cup \{q\}) - v_{\text{MMD}}(\cS) \bigg) \cr
    &= \sum_{r=0}^{d-1} \mu(r) \sum_{\substack{\cS \subseteq \cD \setminus \{q\} \\ |\cS| = r}} \bigg( \frac{1}{n(n-1)} \sum_{i \neq j} \big( k_q(x^{(i)}_q, x^{(j)}_q) - 1 \big) k_{\cS}(\x^{(i)}_{\cS}, \x^{(j)}_{\cS}) \cr
    &\qquad + \frac{1}{m(m-1)} \sum_{i \neq j} \big( k_q(z^{(i)}_q, z^{(j)}_q) - 1 \big) k_{\cS}(\z^{(i)}_{\cS}, \z^{(j)}_{\cS}) \cr
    &\qquad - \frac{2}{nm} \sum_{i, j} \big( k_q(x^{(i)}_q, z^{(j)}_q) - 1 \big) k_{\cS}(\x^{(i)}_{\cS}, \z^{(j)}_{\cS}) \bigg).
\end{align*}
Let \( \cZ^{(\x^{(i)}, \x^{(j)})} = \{ k_1(x^{(i)}_1, x^{(j)}_1), \dots, k_d(x^{(i)}_d, x^{(j)}_d) \} \) and define the elementary symmetric polynomial as
\begin{equation*}
    e_r(\cZ_{-q}^{(\x^{(i)}, \x^{(j)})}) = \sum_{\substack{\cS \subseteq \cD \setminus \{q\} \\ |\cS| = r}} k_{\cS}(\x^{(i)}_{\cS}, \x^{(j)}_{\cS}).
\end{equation*}
Then, it follows from Newton's identities recurrence, with base case $e_0(\cZ_{-q}^{(\x,\x')}) = 1$ and for $r \ge 1$:
\begin{equation*}
    e_r(\cZ_{-q}^{(\x^{(i)}, \x^{(j)})}) = \frac{1}{r} \sum_{s=1}^{r} (-1)^{s-1} e_{r-s}(\cZ_{-q}^{(\x^{(i)}, \x^{(j)})})  p_s(\cZ_{-q}^{(\x^{(i)}, \x^{(j)})}),
\end{equation*}
where \( p_s(\cZ) = \sum_{z \in \cZ} z^s \). Finally, we obtain
\begin{align*}
    \phi_q^{\text{MMD}} &= \frac{1}{n(n-1)} \sum_{i \neq j} \bigg( \big(k_q(x^{(i)}_q, x^{(j)}_q) - 1\big) \sum_{r=0}^{d-1} \mu(r) e_r(\cZ_{-q}^{(\x^{(i)},\x^{(j)})}) \bigg) \cr
    &+ \frac{1}{m(m-1)} \sum_{i \neq j} \bigg( \big(k_q(z^{(i)}_q, z^{(j)}_q) - 1 \big) \sum_{r=0}^{d-1} \mu(r) e_r(\cZ_{-q}^{(\z^{(i)},\z^{(j)})}) \bigg) \cr
    &- \frac{2}{nm} \sum_{i, j} \bigg( \big(k_q(x^{(i)}_q, z^{(j)}_q) - 1 \big) \sum_{r=0}^{d-1} \mu(r) e_r(\cZ_{-q}^{(\x^{(i)},\z^{(j)})}) \bigg),
\end{align*}
and that completes the proof.
\qed

%%%%%%%%%%%%%%%%%%%%%%%%%%%%%%%%
\subsection*{Proof of \Cref{th:hsic func decomp}}
\begin{itemize}[leftmargin=8pt]
    \item[(ii)] We begin by expressing the full kernel matrix as a product of feature-wise kernels:
    \begin{equation*}
        \K = \bigodot_{j \in \cD} \K_j.
    \end{equation*}

    since $\K$ is element-wise product of $\K_j$, we can apply Proposition \ref{th:multiplying kernels} for each element in the kernel matrix and write:
    \begin{equation*}
        \bigodot_{j \in \cD} \K_j = \sum_{\cS \subseteq \cD} \bigodot_{j \in \cS} (\K_j - \onevec\onevec^\top).
    \end{equation*}

    Since HSIC is dependent on \( \mathrm{tr}(\K \bH \bL \bH) \), substituting the decomposition of \( \K \) gives:
    \begin{equation*}
        \widehat{\text{HSIC}}(X, y) =  \frac{1}{(n-1)^2}\mathrm{tr} \left( \bH \bL \bH  \sum_{\cS \subseteq \cD} \bigodot_{j \in \cS} (\K_j - \onevec\onevec^\top) \right).
    \end{equation*}

    Using the linearity of the trace operator, we obtain:
    \begin{equation*}
        \widehat{\text{HSIC}}(X, y) =  \frac{1}{(n-1)^2}\sum_{\cS \subseteq \cD} \mathrm{tr} \!\left( \bH \bL \bH \left[\bigodot_{j \in \cS} (\K_j - \onevec\onevec^\top)\right] \right).
    \end{equation*}

    This completes the proof of the functional decomposition.

    \item[(i)] Following the same reasoning as in Proposition~\ref{prop:func-decomp-value}, the functional decomposition of HSIC in part~(ii) induces a removal operator that sets $k_j=1$ for features in the absent set. The corresponding value function is $v_{\text{HSIC}}(\cS) := P_{\cD\setminus\cS}(\widehat{\text{HSIC}}) - P_{\cD}(\widehat{\text{HSIC}})$. Since $P_{\cD}(\widehat{\text{HSIC}}) = f^{\text{HSIC}}_\emptyset = \frac{1}{(n-1)^2}\mathrm{tr}(\bH\bL\bH\onevec\onevec^\top) = 0$ (as $\bH\onevec=\mathbf{0}$), this reduces to
    \begin{equation*}
        v_{\text{HSIC}}(\cS) = \sum_{\cT \subseteq \cS}  \frac{1}{(n-1)^2}\mathrm{tr} \!\left( \bH \bL \bH \left[\bigodot_{j \in \cT} (\K_j - \onevec\onevec^\top)\right] \right).
    \end{equation*}

    By applying Proposition \ref{th:multiplying kernels}, we simplify:
    \begin{equation*}
        \sum_{\cT \subseteq \cS} \bigodot_{j \in \cT} (\K_j - \onevec\onevec^\top) = \bigodot_{j \in \cS} \K_j.
    \end{equation*}

    Thus, the value function for feature subset \( \cS \) is:
    \begin{equation*}
        v_{\text{HSIC}}(\cS) =  \frac{1}{(n-1)^2}\mathrm{tr} \left( \bH \bL \bH\bigodot_{j \in \cS} \K_j \right) =  \frac{1}{(n-1)^2}\mathrm{tr} \left( \bH \bL \bH \K_{\cS} \right),
    \end{equation*}

    which completes the proof. \qed
\end{itemize}

\subsection*{Proof of Proposition~\ref{prop:hsic sv}}

By substituting  $v_{\text{HSIC}}(\cS) = \frac{1}{(n-1)^2}\mathrm{tr}(\K_{\cS} \bH \bL \bH)$ into the Shapley value formula, we obtain:
\begin{align*}
\phi_j^{\text{HSIC}} &=  \frac{1}{(n-1)^2}\mathrm{tr} \left( \bH \bL \bH\sum_{q=0}^{d-1} \mu(q) \sum_{\substack{\cS \subseteq \cD \setminus \{j\} \\ |\cS| = q}}  \bigg(\K_{\cS\cup \{j\}} - \K_{\cS} \bigg) \right) \cr
&=  \frac{1}{(n-1)^2}\mathrm{tr} \left( \bH \bL \bH\sum_{q=0}^{d-1} \mu(q) \sum_{\substack{\cS \subseteq \cD \setminus \{j\} \\ |\cS| = q}} \bigg(\K_j \odot \K_{\cS} - \K_{\cS} \bigg) \right) \cr
&=  \frac{1}{(n-1)^2} \mathrm{tr} \left( \bH \bL \bH(\K_j - \onevec\onevec^\top) \odot \sum_{q=0}^{d-1} \mu(q) \sum_{\substack{\cS \subseteq \cD \setminus \{j\} \\ |\cS| = q}} \K_{\cS} \right).
\end{align*}

Letting $\cK = \{\K_1, ..., \K_d\}$, we express:
\[
\sum_{\substack{\cS \subseteq \cD \setminus \{j\} \\ |\cS| = q}}  \K_{\cS} = E_q(\cK_{-j}),
\]
which is computed recursively via Newton's identities, with base case $E_0(\cK_{-j}) = \onevec\onevec^\top$ and for $q \ge 1$:
\[
E_q(\cK_{-j}) = \frac{1}{q} \sum_{r=1}^{q} (-1)^{r-1} E_{q-r}(\cK_{-j}) \odot P_r(\cK_{-j}),
\]
where $P_r(\cK) = \sum_{\K_i \in \cK} \K_i^{\odot r}$ is the element-wise power sum (with $\K_i^{\odot r}$ denoting the element-wise $r$-th power of $\K_i$). Substituting this back, we obtain:
\[
\phi_j^{\text{HSIC}} =  \frac{1}{(n-1)^2}\mathrm{tr} \left( \bH \bL \bH\bigg( (\K_j - \onevec\onevec^\top) \bigodot \sum_{q=0}^{d-1} \mu(q) E_q(\cK_{-j}) \bigg) \right),
\]
which completes the proof. \qed

%%%%%%%%%%%%%%%%%%%%%%%%%%%%%%%%%%%%%%%%%%%%%%%%%%%%%%%%%%%%
%%%%%%%%%%%% Interpolation-Based ESP Recovery Methods
%%%%%%%%%%%%%%%%%%%%%%%%%%%%%%%%%%%%%%%%%%%%%%%%%%%%%%%%%%%%
\section{Interpolation-Based ESP Recovery Methods}
\label{apx:interpolation}

This appendix describes the polynomial interpolation approaches used to recover the elementary symmetric polynomials (ESPs) appearing in the recursive Shapley formulation of Theorem~\ref{th:shapley recursive}. Both the Chebyshev and Fourier constructions encode the ESPs
\[
\{e_q(\cZ_{-j})\}_{q=0}^{d-1},
\qquad j \in \{1,\dots,d\},
\]
as the coefficients of a univariate polynomial and attempt to reconstruct these coefficients from evaluations of the polynomial at a fixed set of interpolation nodes. Although highly effective for tree-based models---where the relevant polynomial degree is the depth of the tree---these methods become numerically unstable for product kernels, where the polynomial degree equals $d-1$, the total number of features. We describe each method in detail and analyze its numerical behavior.

\subsection{Chebyshev Interpolation}
\label{apx:chebyshev}

\paragraph{Polynomial Representation.}
Theorem~\ref{th:shapley recursive} shows that the Shapley value for feature $j$ involves the weighted sum
\[
\sum_{q=0}^{d-1} \mu(q)\, e_q(\cZ_{-j}),
\]
where $e_q(\cZ_{-j})$ is the degree-$q$ ESP over the leave-one-out set $\cZ_{-j} = \cZ \setminus \{\z_j\}$. Define the polynomial
\[
H_j(y)
\;:=\;
\sum_{q=0}^{D} e_q(\cZ_{-j})\, y^q,
\qquad D := d-1.
\]
Recovering the ESPs $\{e_q(\cZ_{-j})\}$ is therefore equivalent to recovering the coefficient vector of $H_j$.

\paragraph{Interpolation at Chebyshev Nodes.}
Let
\[
y_\ell = \cos\!\left(\frac{\pi \ell}{D}\right), 
\qquad \ell = 0,\dots,D,
\]
be the $(D+1)$ Chebyshev points of the second kind on $[-1,1]$. Evaluating the polynomial at these nodes yields
\[
g_{j,\ell}
=
H_j(y_\ell)
=
\sum_{q=0}^{D} e_q(\cZ_{-j})\, y_\ell^q.
\]
Define the vector $e^{(-j)} = (e_0(\cZ_{-j}),\dots,e_D(\cZ_{-j}))^\top$ and the Chebyshev--Vandermonde matrix $V \in \mathbb{R}^{(D+1)\times(D+1)}$ with entries
\[
V_{\ell q} = y_\ell^q.
\]
Then
\[
g_j = V\, e^{(-j)}.
\]
Inverting this system recovers $e^{(-j)} = V^{-1} g_j$. In practice, instead of repeatedly solving this system, it is convenient to precompute a weight vector $N \in \mathbb{R}^{D+1}$ satisfying
\[
\mu^\top e^{(-j)} = N^\top g_j,
\]
so that the Shapley contribution can be evaluated without explicitly materializing $e^{(-j)}$.

\paragraph{Numerical Instability.}
Although optimal in an approximation-theoretic sense, Chebyshev nodes do not prevent numerical ill-conditioning when the polynomial degree $D$ is moderate or large. Even with Chebyshev spacing, the condition number of $V$ grows exponentially with $D$: for $D\approx 40$, one typically obtains $\mathrm{cond}(V) \gtrsim 10^{12}$ in double precision, and for $D\approx 60$ the matrix becomes nearly singular. In decision trees, the polynomial degree equals the depth, often $<15$. For product kernels, the degree is $D = d-1$, which can be up to hundreds, making instability intrinsic. These issues make the inversion $V^{-1} g_j$ highly unreliable beyond $d \approx 40$, even if the Chebyshev grid is used.

\subsection{Fourier / Roots-of-Unity Interpolation}
\label{apx:fourier}

The Fourier approach mirrors the Chebyshev strategy but replaces Chebyshev nodes with equally spaced complex roots of unity. The algebraic structure becomes particularly simple due to the unitary nature of the discrete Fourier transform (DFT).

\paragraph{Roots-of-Unity Evaluation.}
Let $n = D+1$ and
\[
\omega_\ell = \exp\!\left(\tfrac{2\pi i \ell}{n}\right),
\qquad \ell = 0,\dots,n-1.
\]
Evaluating $H_j$ at these points gives
\[
g_{j,\ell}
=
H_j(\omega_\ell)
=
\sum_{q=0}^D e_q(\cZ_{-j}) \, \omega_\ell^q.
\]
Define the DFT matrix $F \in \mathbb{C}^{n\times n}$ by $F_{\ell q} = \omega_\ell^q$. Then
\[
g_j = F\, e^{(-j)}, 
\qquad
e^{(-j)} = F^{-1} g_j = \frac{1}{n} F^\ast g_j.
\]

\paragraph{Relation to the Chebyshev Construction.}
The Fourier method uses the same polynomial representation as the Chebyshev method, but replaces the real Vandermonde matrix $V$ with the complex DFT matrix $F$. The main steps differ only in the choice of nodes and the linear system that is solved:
\[
\text{Chebyshev:}\quad g_j = V e^{(-j)},
\qquad
\text{Fourier:}\quad g_j = F e^{(-j)}.
\]
Both methods require evaluating $H_j$ at $D+1$ distinct interpolation points and both require recovering the coefficients of a degree-$D$ polynomial from these evaluations.

\paragraph{Numerical Instability.}
Although the DFT matrix $F$ is perfectly conditioned in exact arithmetic, numerical instability arises for product kernels due to various reasons. First and foremost, the values $H_j(\omega_\ell)$ involve complex products $\prod_{i\neq j}(1+\omega_\ell \z_i)$ whose magnitudes grow roughly exponentially in $d$. Thus $g_j$ spans an enormous dynamic range. In addition, the inverse DFT involves alternating complex phases. When some entries of $g_j$ are extremely large and others small, cancellation magnifies floating-point error. Moreover, as with the Chebyshev method, the instability is structural: the degree $D=d-1$ is simply too large for stable recovery of polynomial coefficients by interpolation.

Even though the Fourier transform itself is numerically stable, the interpolation problem it is solving is not. In our experiments, instability appears as early as $d \approx 50$ for product kernels.

\subsection{Numerical Stability of ESP-Based Shapley Computation}
\label{subsec:esp_stability}

We empirically evaluate the numerical stability and computational efficiency of different elementary symmetric polynomial (ESP) aggregation strategies used to compute exact Shapley values in product-kernel models.

\paragraph{Experimental setup.}
Synthetic regression datasets are generated using \texttt{sklearn.make\_regression} with $n=500$ samples and varying feature dimension
$d \in \{10, 15, 20, 25, 30, 40, 50, 70, 80, 100, 200, 500, 1000\}$.
For each $d$, the data are split into $90\%$ training and $10\%$ test sets.
A Gaussian Process Regressor with an RBF kernel is trained on the training set, using a small observation noise term ($\alpha=10^{-2}$) and no hyperparameter restarts unless otherwise stated.

Local explanations are computed using a product-kernel Shapley explainer, instantiated with three different ESP aggregation strategies:
(i) \emph{quadratic}, a dynamic-programming method based on prefix--suffix recursion over elementary symmetric polynomials;
(ii) \emph{FFT}, which aggregates ESPs via roots-of-unity evaluation and fast Fourier transforms;
and (iii) \emph{Chebyshev}, which interpolates the ESP polynomial using Chebyshev nodes and solves the resulting Vandermonde system.

\paragraph{Evaluation protocol.}
For each trained model and each ESP method, we randomly select five test instances and compute per-feature Shapley values.
Correctness is assessed using the Shapley efficiency identity
\begin{equation}
    \sum_{j=1}^d \phi_j \;=\; f(x) - v(\emptyset),
\end{equation}
where $f(x)$ is the GP prediction and $v(\emptyset)$ denotes the null-game value used by the explainer.
We report the \emph{average absolute Shapley-sum error},
$\mathbb{E}\bigl[\,\lvert \sum_j \phi_j - (f(x) - v(\emptyset)) \rvert \,\bigr]$,
as well as the average wall-clock time per explanation call.
All results are averaged over the five trials.

\paragraph{Results.}
Table~\ref{tab:esp_benchmark} reports numerical error and runtime across feature dimensions.
The quadratic dynamic-programming method exhibits consistently low numerical error, on the order of $10^{-13}$--$10^{-12}$, across all tested dimensions, including $d=1000$.
Runtime increases smoothly with $d$, remaining below one second up to $d=500$ and around $2.6$ seconds at $d=1000$.

In contrast, both FFT- and Chebyshev-based approaches are accurate for small to moderate dimensions ($d \lesssim 20$--$25$), achieving near machine-precision error, but exhibit rapid numerical degradation as $d$ increases.
Beyond this regime, errors grow exponentially and quickly become catastrophic, despite these methods maintaining low wall-clock times.

\begin{table}[t]
\centering
\caption{Numerical accuracy and runtime of ESP-based Shapley computation.
We report the average absolute Shapley-sum error
$\mathbb{E}[|\sum_j \phi_j - (f(x)-v(\emptyset))|]$
and average wall-clock time per explanation call (seconds), averaged over 5 trials.}
\label{tab:esp_benchmark}
\small
\begin{tabular}{r rr rr rr}
\toprule
$d$
& \multicolumn{2}{c}{PKeX-Shapley}
& \multicolumn{2}{c}{FFT}
& \multicolumn{2}{c}{Chebyshev} \\
\cmidrule(lr){2-3}
\cmidrule(lr){4-5}
\cmidrule(lr){6-7}
& Error & Time (s)
& Error & Time (s)
& Error & Time (s) \\
\midrule
10   & $2.26{\times}10^{-12}$ & $1.22{\times}10^{-3}$
     & $4.83{\times}10^{-12}$ & $1.55{\times}10^{-3}$
     & $4.35{\times}10^{-10}$ & $1.38{\times}10^{-3}$ \\
15   & $2.34{\times}10^{-12}$ & $2.10{\times}10^{-3}$
     & $2.07{\times}10^{-10}$ & $2.97{\times}10^{-3}$
     & $4.86{\times}10^{-7}$  & $2.28{\times}10^{-3}$ \\
20   & $1.14{\times}10^{-12}$ & $2.88{\times}10^{-3}$
     & $2.07{\times}10^{-9}$  & $4.13{\times}10^{-3}$
     & $2.61{\times}10^{-4}$  & $2.71{\times}10^{-3}$ \\
25   & $7.16{\times}10^{-13}$ & $3.75{\times}10^{-3}$
     & $5.05{\times}10^{-8}$  & $5.04{\times}10^{-3}$
     & $2.29{\times}10^{-1}$  & $3.67{\times}10^{-3}$ \\
30   & $1.01{\times}10^{-12}$ & $4.78{\times}10^{-3}$
     & $1.03{\times}10^{-6}$  & $6.94{\times}10^{-3}$
     & $1.70{\times}10^{2}$   & $4.52{\times}10^{-3}$ \\
40   & $1.93{\times}10^{-13}$ & $7.36{\times}10^{-3}$
     & $7.43{\times}10^{-4}$  & $9.93{\times}10^{-3}$
     & $4.59{\times}10^{7}$   & $7.02{\times}10^{-3}$ \\
50   & $3.64{\times}10^{-13}$ & $1.05{\times}10^{-2}$
     & $9.77{\times}10^{-3}$  & $1.68{\times}10^{-2}$
     & $2.77{\times}10^{10}$  & $9.89{\times}10^{-3}$ \\
70   & $4.55{\times}10^{-13}$ & $1.81{\times}10^{-2}$
     & $8.15{\times}10^{0}$   & $2.97{\times}10^{-2}$
     & $2.99{\times}10^{16}$  & $1.79{\times}10^{-2}$ \\
80   & $7.28{\times}10^{-13}$ & $2.23{\times}10^{-2}$
     & $1.59{\times}10^{4}$   & $3.66{\times}10^{-2}$
     & $3.08{\times}10^{18}$  & $2.20{\times}10^{-2}$ \\
100  & $1.07{\times}10^{-12}$ & $3.38{\times}10^{-2}$
     & $5.28{\times}10^{7}$   & $5.55{\times}10^{-2}$
     & $1.53{\times}10^{22}$  & $3.20{\times}10^{-2}$ \\
200  & $5.46{\times}10^{-13}$ & $1.18{\times}10^{-1}$
     & $8.60{\times}10^{27}$  & $2.19{\times}10^{-1}$
     & $3.93{\times}10^{45}$  & $1.15{\times}10^{-1}$ \\
500  & $2.50{\times}10^{-13}$ & $6.98{\times}10^{-1}$
     & $3.57{\times}10^{85}$  & $1.71{\times}10^{0}$
     & $1.10{\times}10^{102}$ & $9.61{\times}10^{-1}$ \\
1000 & $1.73{\times}10^{-12}$ & $2.65{\times}10^{0}$
     & $2.90{\times}10^{183}$ & $1.36{\times}10^{1}$
     & $3.72{\times}10^{199}$ & $3.75{\times}10^{0}$ \\
\bottomrule
\end{tabular}
\end{table}

\paragraph{Discussion.}
These results highlight a clear trade-off between computational speed and numerical robustness.
While FFT- and Chebyshev-based ESP aggregation can be attractive for low-dimensional problems, their current formulations amplify rounding and conditioning errors as the polynomial degree grows, leading to numerical instability at moderate to large $d$.
The quadratic dynamic-programming approach, although asymptotically slower, is the only method that remains numerically reliable across all tested dimensions, making it the appropriate default for exact Shapley computation in high-dimensional product-kernel models.

%%%%%%%%%%%%%%%%%%%%%%%%%%%%%%%%%%%%%%%%%%%%%%%%%%%%%%%%%%%%
% % %%%%%%%%%%%% Experiments
% % %%%%%%%%%%%%%%%%%%%%%%%%%%%%%%%%%%%%%%%%%%%%%%%%%%%%%%%%%%%%
\section{Experiments}\label{apx:experiments}

\subsection{Experimental Setup}\label{apx:experimental setup}
\paragraph{SVM Optimization Using Optuna \citep{optuna}} When using SVM in our experiments, we optimized the Support Vector Machine (SVM) with a Radial Basis Function (RBF) kernel using Optuna~\citep{optuna}, a robust hyperparameter optimization framework. The target type (either 'regression' or 'classification') was determined to guide the selection of the appropriate SVM model (\texttt{SVR} for regression and \texttt{SVC} for classification). The hyperparameters \texttt{C} and \texttt{gamma}, critical for the RBF kernel's performance, were optimized within an extensive range using a log-uniform distribution. Specifically, we defined the hyperparameters: \texttt{C} between $10^{-5}$ and $10^{5}$ and \texttt{gamma} between $10^{-5}$ and $10^{3}$, and utilized 5-fold cross-validation to ensure reliable evaluation. The optimization process aimed to minimize the mean squared error for regression tasks and maximize accuracy for classification tasks. After conducting the specified number of trials (n=100), the best hyperparameters were used to train a final SVM model on the entire dataset, yielding both the optimal model configuration and the best cross-validation score achieved during the optimization process.

\paragraph{Training Gaussian Process (GP) Using K-Fold Cross-Validation}
When using GP, we trained a model using k-fold cross-validation to ensure robust evaluation and generalization performance. We defined the GP kernel as \texttt{C(1.0, (1e-4, 1e1)) * RBF(1.0, (1e-4, 10))}, suitable for both regression and classification tasks. For classification problems, {GaussianProcessClassifier} was utilized, while \texttt{GaussianProcessRegressor} was used for regression tasks. We employed \texttt{K-Fold} with $K=5$ for cross-validation to evaluate the model's performance across different folds. All the hyperparameters, including the kernel width of the RBF kernel, are determined in the training process using optimization. During the cross-validation process, the model was trained on each fold, and predictions were made on the validation fold. Performance metrics were chosen based on the problem type: accuracy for classification models and mean absolute percentage error (MAPE) for regression models. The scores from each fold were aggregated to compute the average and standard deviation of the scores.

% \subsection{Execution time}\label{apx:exp computation}
% To assess the computational efficiency of our recursive algorithm, we conducted a simulation study using a randomly generated kernel function of the form $\alphab^\top k(\x, \X)$ with 1000 samples. We computed Shapley values under both the brute-force enumeration and our recursive method by varying the number of features, as plotted in Figure \ref{fig:computation time}. With a runtime budget of 300 seconds, the brute-force computation was feasible only for up to 30 features, already requiring more than 200 seconds at this scale, while exceeding the budget beyond that. In contrast, our recursive algorithm consistently completed the same computations in a fraction of the time, from milliseconds at lower dimensions to under 100 seconds, even with 500 features. These results clearly highlight the substantial efficiency gains and scalability of our method for high-dimensional settings.

\subsection{MMD Experiments}\label{apx:mmd exp}
\begin{figure}
    \centering
    \includegraphics[width=\linewidth]{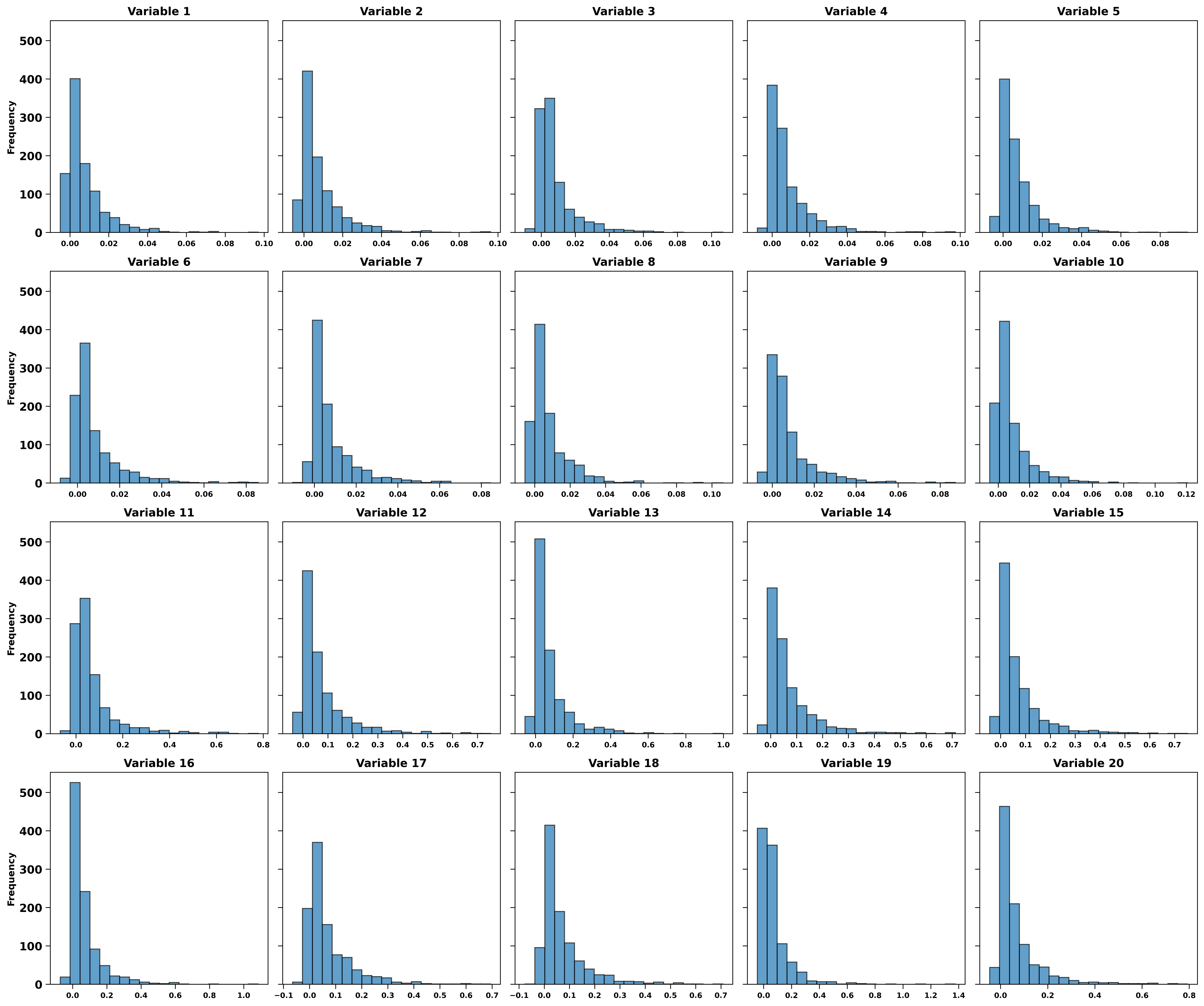}
    \caption{Shapley values for the synthetic data sets with equal distributions. All variables contribute equally to the near-zero MMD.}
    \label{fig:mmd case2}
\end{figure}

In addition to the synthetic experiments for MMD, we first provide another experiment for the cases when there is no distribution discrepancy. To that end, $X$ and $Z$ are sampled from the same multivariate normal distribution across all 20 variables. The MMD is near zero, indicating the distributions are equivalent. Shapley values are computed and replicated 1000 times, with histograms plotted for each variable in Figure \ref{fig:mmd case2}. The near-identical distributions of Shapley values across all variables reflect the uniform contribution of these variables to the MMD close to zero, consistent with the absence of any significant difference between the distributions.

We extend our analysis to the UCI Diabetes dataset, consisting of 442 samples and 10 baseline variables, including age, sex, body mass index (BMI), average blood pressure, and six blood serum measurements (shown by s1 to s6 features). The dataset is split into male and female subsets using the second variable (sex), which is excluded from the analysis, leaving nine variables for comparison.

\begin{figure}
    \centering
    \includegraphics[width=0.8\linewidth]{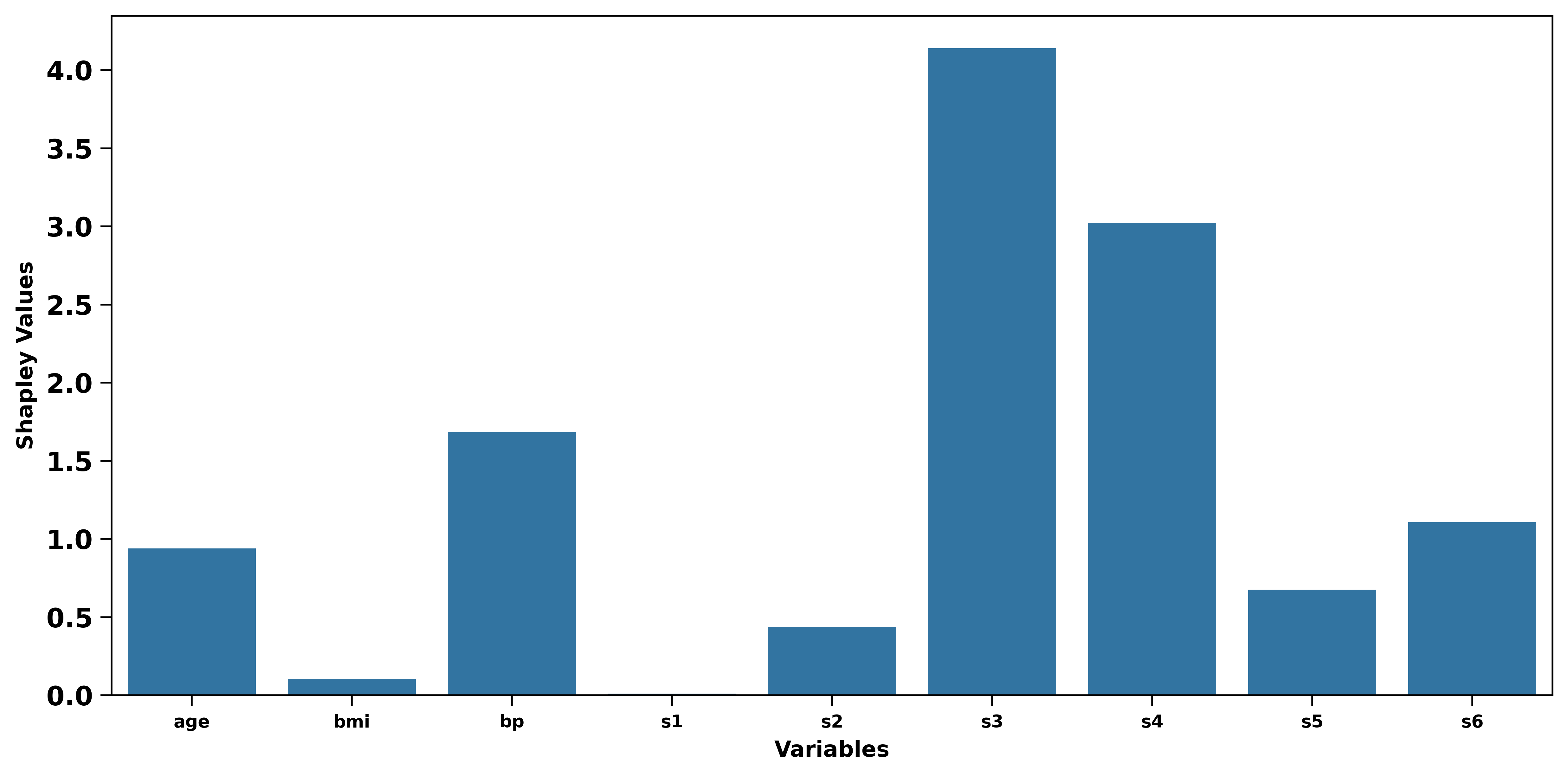}
    \caption{Shapley values explaining MMD between male and female subsets in the UCI Diabetes data set.}

    \label{fig:mmd diabetes sv}
\end{figure}

Using MMD, we calculate the dissimilarity between male and female groups and then compute Shapley values to attribute variable contributions to the MMD. Figure~\ref{fig:mmd diabetes sv} displays the Shapley values for the nine variables in the Diabetes dataset. The results show that $s3$ and $s4$ contribute most significantly to the MMD, followed by $bp$, $s6$, $age$, $s5$, and $s2$. In contrast, $bmi$ and $s1$ contribute nearly zero to the MMD, indicating their alignment across the two groups. 

\begin{figure}
    \centering
    \includegraphics[width=\linewidth]{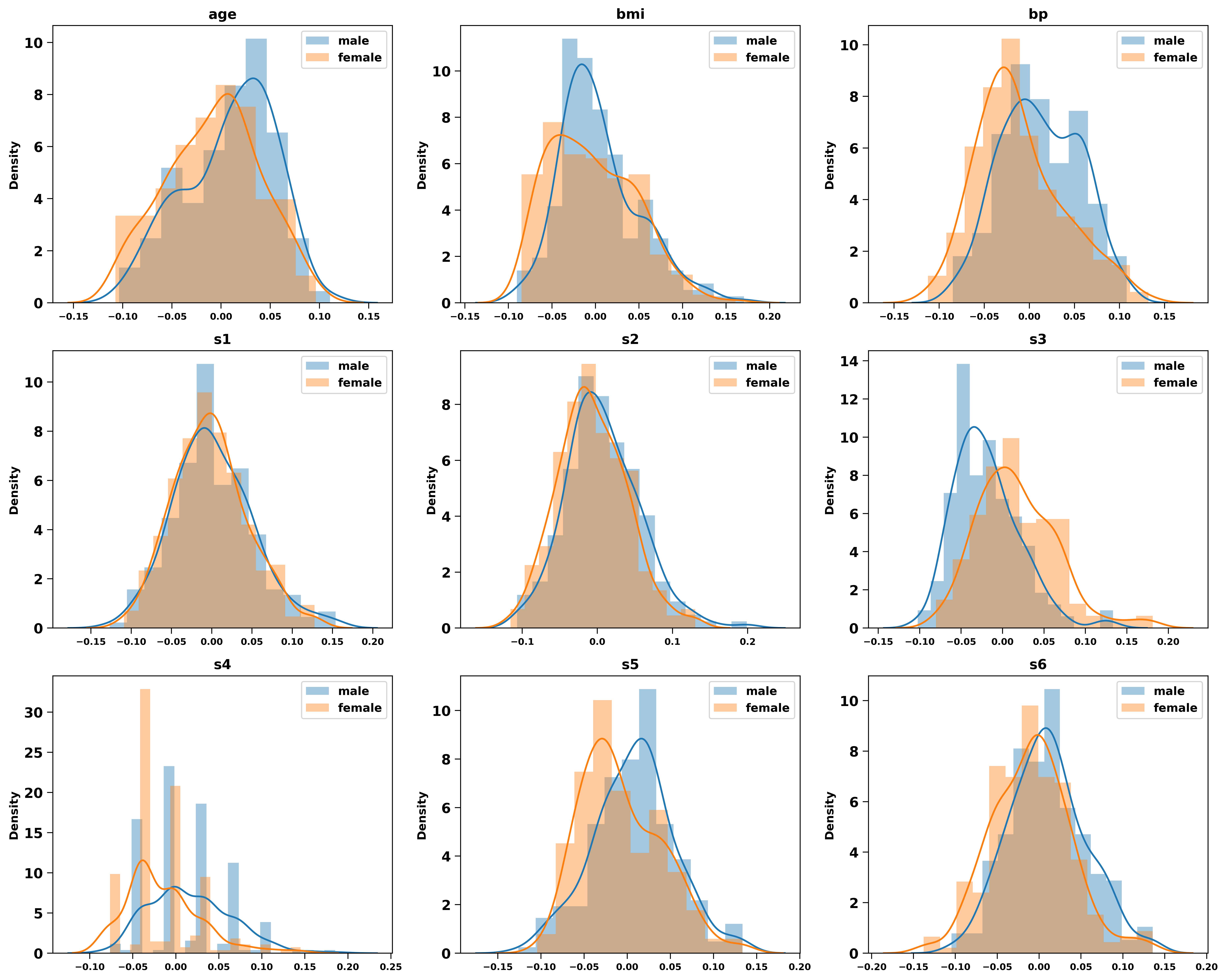}
    \caption{Marginal distributions of variables for male and female subsets in the UCI Diabetes dataset.}

    \label{fig:mmd diabetes marginals}
\end{figure}

To validate these results, we analyze the marginal distributions of variables for males and females, as shown in Figure~\ref{fig:mmd diabetes marginals}. The analysis confirms that variables with distinct marginal distributions between males and females (e.g., $s3$ and $s4$) have high positive Shapley values, reflecting their role in increasing the MMD. Conversely, variables with similar distributions (e.g., $s1$) exhibit zero or infinitesimal Shapley values, implying they do not contribute to the overall MMD and the discrepancy between the two distributions.

\subsection{HSIC Feature Selection Case Study}\label{apx:hsic exp}

We demonstrate how attributing the HSIC between input features and the target variable to individual features can support feature selection by quantifying their respective contributions to the overall statistical dependence. We compare our approach with five feature importance methods: HSICLasso~\citep{hsiclasso}, Mutual Information (MI), Lasso, K–Best, and Tree Ensemble (with feature permutations). Experiments were conducted on seven datasets, where we trained Gaussian Process~(GP) models with an RBF kernel, using only the top 20\% of features ranked by each selection method.

Table~\ref{tab:hsic-20pct} reports the five‐fold cross‐validated mean and standard deviation for the GP models trained on the selected features. We use mean absolute percentage error (MAPE) for regression tasks and accuracy for classification tasks. For kernel computation in HSIC, we use an RBF kernel for features and regression targets, and a categorical kernel for classification targets, with bandwidth selected via the median heuristic. PKeX-Shapley consistently delivers strong results across all datasets. On regression problems, it yields better MAPE on \textit{breast cancer}, \textit{skillcraft}, and \textit{parkinson}, while other methods often incur higher error or greater variance. For classification, PKeX-Shapley maintains accuracies above 80\% (except for one case), matching or exceeding the baselines. Notably, it outperforms HSICLasso---a method specifically tailored for feature selection---on 5 out of 7 datasets, tying on one and only falling short on \textit{ionosphere}.

\begin{table*}[!h]
\centering
\footnotesize
\setlength{\tabcolsep}{2.5pt}
\caption{Performance (mean±standard deviation) when training on the top 20\% of features.
Datasets \textit{breast cancer}, \textit{skillcraft}, \textit{sml}, and \textit{parkinson} are regression (MAPE, lower is better);
\textit{sonar}, \textit{Wisconsin}, \textit{ionosphere} are classification (accuracy, higher is better).}
\vspace{-.5em}
\label{tab:hsic-20pct}
\resizebox{\textwidth}{!}{
\begin{tabular}{lccc|cccc}
\toprule
Method & \textbf{sonar} & \textbf{Wisconsin} & \textbf{ionosphere}
& \textbf{breast cancer} & \textbf{skillcraft} & \textbf{sml} & \textbf{parkinson} \\
\midrule
& \multicolumn{3}{c|}{accuracy ($\uparrow$)}
& \multicolumn{4}{c}{mean absolute percentage error ($\downarrow$)} \\
\midrule\midrule
PKeX-Shapley
& ${0.808}_{\pm 0.030}$
& $\mathbf{{0.909}_{\pm 0.015}}$
& ${0.878}_{\pm 0.036}$
& $\mathbf{{0.850}_{\pm 0.010}}$
& $\mathbf{{1.000}_{\pm 0.020}}$
& ${0.999}_{\pm 0.001}$
& $\mathbf{{0.125}_{\pm 0.022}}$ \\

HSICLasso
& ${0.808}_{\pm 0.044}$
& ${0.884}_{\pm 0.010}$
& ${0.912}_{\pm 0.035}$
& ${1.000}_{\pm 0.080}$
& ${2.175}_{\pm 0.429}$
& ${1.354}_{\pm 0.726}$
& ${1.170}_{\pm 0.269}$ \\

MI
& $\mathbf{{0.875}_{\pm 0.053}}$
& ${0.900}_{\pm 0.015}$
& $\mathbf{{0.937}_{\pm 0.023}}$
& ${1.000}_{\pm 0.080}$
& ${1.134}_{\pm 0.099}$
& $\mathbf{{0.196}_{\pm 0.071}}$
& ${0.214}_{\pm 0.004}$ \\

Lasso
& ${0.842}_{\pm 0.038}$
& ${0.900}_{\pm 0.024}$
& ${0.932}_{\pm 0.021}$
& ${1.000}_{\pm 0.080}$
& ${1.821}_{\pm 0.680}$
& ${1.000}_{\pm 0.000}$
& ${1.000}_{\pm 0.000}$ \\

K--Best
& ${0.779}_{\pm 0.040}$
& $\mathbf{{0.909}_{\pm 0.015}}$
& ${0.869}_{\pm 0.018}$
& ${1.000}_{\pm 0.080}$
& ${1.134}_{\pm 0.099}$
& ${0.257}_{\pm 0.086}$
& ${1.018}_{\pm 0.076}$ \\

Tree Ens.
& ${0.837}_{\pm 0.059}$
& ${0.887}_{\pm 0.033}$
& ${0.926}_{\pm 0.031}$
& ${1.000}_{\pm 0.080}$
& ${2.175}_{\pm 0.429}$
& ${1.000}_{\pm 0.000}$
& ${0.214}_{\pm 0.004}$ \\

\bottomrule
\end{tabular}}
\end{table*}

\section{FAQ}\label{apx:faq}
\subsection*{How Does PKeX-Shapley Differ from \citet{mohammadi2025exact}?}
In \citet{mohammadi2025exact}, the authors propose an exact Shapley computation method for stochastic attribution in FANOVA GP models. Similar to PKeX-Shapley, their approach leverages Newton’s identities and symmetric polynomial representations to obtain closed-form attributions. However, the two methods differ in both scope and capability.

PKeX-Shapley is designed specifically for kernel methods with product kernels. It employs the functional decomposition value function to compute exact Shapley values. While this formulation enables exact computation of the mean Shapley values, it does not extend naturally to higher-order moments. In particular, when applied to Gaussian processes, computing the variance or covariance structure of Shapley values is nontrivial, since orthogonality does not generally hold for product kernels. As a result, our method cannot be directly extended to explain GP models in a stochastic way. In addition, PKeX-Shapley computes all d Shapley values in $O(d^2n)$, while the one proposed by \citet{mohammadi2025exact} has the same complexity for compute one Shapley value only. 

In addition, \citet{mohammadi2025exact} focus on FANOVA GPs, whose kernels admit a functional ANOVA decomposition. Compared to product kernels, this class of kernels is more restrictive and often harder to train in practice. Nonetheless, the additional structure provides a key advantage: it enables exact polynomial-time computation not only of the mean but also of the variance and covariance of stochastic Shapley values—something that product kernels cannot achieve. This distinction highlights the central difference between their method and ours.

% In contrast, their method is built on top of FANOVA GPs, where the underlying kernel has a functional ANOVA decomposition. This decomposition enables us to define a more expressive value function that captures the marginal contribution of each feature subset. However, computing Shapley values directly from this value function is intractable due to its complexity. To address this, the authors introduce a (stochastic) mobius representation of the Shapley value, which allows us to compute exact means and variances of Shapley values using recursive and efficient algorithms. 

\subsection*{How Does PKeX-Shapley Differ from RKHS-SHAP~\citep{rkhs_shap}?}

RKHS-SHAP is the first kernel method-specific SHAP-based algorithm. While the author still employs the conditional expectation value function ${\tilde{\nu}_{\x}}(S) = \mathbb{E}[f(X)\mid X_S=\x_S]$, they used the fact that for function $f$ in the RKHS $\mathcal{H}_k$, it is possible to estimate ${\tilde{\nu}_{\x}}(S)$ non-parametrically utilizing a tool known as conditional kernel mean embedding. Specifically, this leads to the following expression:
\begin{align*}
    {\tilde{\nu}_{\x}}(\cS) = \boldsymbol{\alpha}^\top \K(\K_{\cS} + n\lambda I)^{-1}k_{\cS}(\X_{\cS}, \x_{\cS})
\end{align*}
which of course, is different from our functional decomposition value function.

Another crucial difference between PkeX-Shapley and \citet{rkhs_shap} is that the former uses two recurrences to compute all Shapley values in quadratic time, while the latter uses a subset of coalition samples and computes the Shapley value using a regressor, similar to Kernel SHAP.

% Although looking different in their expression, there is an interesting mathematical connection that is not apparent at first glance. First of all, we need to understand that conditional expectations can be interpreted as orthogonal projections to the space of nice-behaving functions defined on the conditioning variable. Specifically, let $\mathcal{L}^2(\mathcal{X})$ be the space of square-integrable functions on $X$, then for a random variable $Y$, the conditional expectation $\mathbb{E}[Y\mid X]$ is the unique element in $\mathcal{L}^2(\mathcal{X})$ that is closest to $Y$ in mean-square error:
% \begin{align*}
%     \mathbb{E}[Y\mid X] = \arg \min_{Z\in \mathcal{L}^2(\mathcal{X})}\mathbb{E}[(Y-Z)^2].
% \end{align*}

% As a result, this interpretation of conditional expectations, and thus of the conditional-expectation-based value function \(\tilde{\nu}_{\x}\), allows us to directly connect it to our functional baseline value function. As established in Proposition~\ref{prop: orthogonal_projection} and Appendix~\ref{apx:orthogonal projection}, the latter can also be viewed as an orthogonal projection, albeit onto a different function space.

\end{document}